\documentclass[pdflatex,sn-apa]{sn-jnl}% APA Reference Style 
\usepackage{graphicx}%
\usepackage{multirow}%
\usepackage{amsmath,amssymb,amsfonts}%
\usepackage{amsthm}%
\usepackage{mathrsfs}%
\usepackage[title]{appendix}%
\usepackage{xcolor}%
\usepackage{textcomp}%
\usepackage{manyfoot}%
\usepackage{booktabs}%
\usepackage{algorithm}%
\usepackage{algorithmicx}%
\usepackage{algpseudocode}%
\usepackage{listings}%
\usepackage{siunitx}
\usepackage{tcolorbox}
\usepackage{hyperref}
\usepackage{float}

\usepackage{graphicx}
\usepackage[maxfloats=256]{morefloats}
\usepackage{subcaption}

\usepackage{threeparttable}
\usepackage{booktabs}

\usepackage{fancyvrb}

\usepackage{xcolor,pifont}
\newcommand*\colourcheck[1]{%
	\expandafter\newcommand\csname #1check\endcsname{\textcolor{#1}{\ding{52}}}%
}
\newcommand*\colourcross[1]{%
	\expandafter\newcommand\csname #1cross\endcsname{\textcolor{#1}{\ding{56}}}%
}
\colourcheck{green}
\colourcross{red}

\begin{document}

\title[SweCTRL-Mini: a data-transparent Transformer-based large language model for controllable text generation in Swedish]{SweCTRL-Mini: a data-transparent Transformer-based large language model for controllable text generation in Swedish}


\author*[1]{\fnm{Dmytro} \sur{Kalpakchi}}\email{dmytroka@kth.se}

\author[1]{\fnm{Johan} \sur{Boye}}\email{jboye@kth.se}

\affil*[1]{\orgdiv{Division of Speech, Music and Hearing}, \orgname{KTH Royal Institute of Technology}, \orgaddress{\street{Lindstedtsvägen 24}, \city{Stockholm}, \postcode{10044}, \country{Sweden}}}


\abstract{We present SweCTRL-Mini, a large Swedish language model that can be used for inference and fine-tuning on a single consumer-grade GPU. The model is based on the CTRL architecture by \cite{keskar2019ctrl}, which means that users of the SweCTRL-Mini model can control the genre of the generated text by inserting special tokens in the generation prompts. SweCTRL-Mini is trained on a subset of the Swedish part of the mC4 corpus and a set of Swedish novels. In this article, we provide (1) a detailed account of the utilized training data and text pre-processing steps, to the extent that it is possible to check whether a specific phrase/source was a part of the training data, and (2) an evaluation of the model on both discriminative tasks, using automatic evaluation methods, and generative tasks, using human referees. We also compare the generative capabilities of the model with those of GPT-3. SweCTRL-Mini is fully open and available for download.}

%%================================%%
%% Sample for structured abstract %%
%%================================%%

% \abstract{\textbf{Purpose:} The abstract serves both as a general introduction to the topic and as a brief, non-technical summary of the main results and their implications. The abstract must not include subheadings (unless expressly permitted in the journal's Instructions to Authors), equations or citations. As a guide the abstract should not exceed 200 words. Most journals do not set a hard limit however authors are advised to check the author instructions for the journal they are submitting to.




\maketitle

\section{Introduction}
Models based on a Transformer architecture \citep{vaswani2017attention} have made a substantial impact in Natural Language Processing (NLP), especially for English, and are being used for many discriminative and generative tasks \citep{LIN2022111,tay2022efficient}. In this article, we are in particular interested in models capable of solving generative tasks. While the landscape of such models becomes broader for Swedish (as detailed in Section \ref{sec:relwork}), the existing models lack two things:
\begin{itemize}
	\item a detailed account of the utilized training data and text pre-processing steps, to the extent that it is possible to check whether a specific phrase/source was a part of the training data and if so, in what context;
	\item an evaluation of \emph{generative} models on \emph{generative} tasks, going beyond automatic evaluation metrics.
\end{itemize}
Furthermore, the trend in Swedish NLP (and in NLP in general) is to make generative models larger and larger, to the extent that inference is impossible on a single GPU\footnote{This seems to be the case with most of the yet unpublished GPT-SW3 models \url{https://www.ai.se/en/node/81535/gpt-sw3}}.

In this article we attempt to provide a fresh perspective and introduce SweCTRL-Mini, a Transformer-based model for Swedish, allowing \textbf{both inference and fine-tuning on a single GPU}! On top of that, we fully disclose our training data and exact pre-processing steps, and provide an interface for searching 13-grams of our model's training data, and URLs for pages included in the training data\footnote{Guaranteed to be available online until 30 April 2024, the scripts to host such instance on your own premises are available at the associated GitHub repository.}. The links to the source code, the SweCTRL-Mini model, and all accompanying resources are available at \url{https://github.com/dkalpakchi/SweCTRL-Mini}.

\section{Related work}
\label{sec:relwork}
In recent years, the Swedish NLP community has produced a number of Transformer-based models. Those aimed at solving the discriminative tasks include BERT-based models, more specifically:
\begin{itemize}
	\item \cite{swedish-bert} produced Swedish versions of BERT \citep{devlin-etal-2019-bert}, ALBERT \citep{lan2019albert}, and ELECTRA \citep{clark2020electra};
	\item Swedish versions\footnote{\url{https://huggingface.co/flax-community/swe-roberta-wiki-oscar}}\footnote{\url{https://huggingface.co/birgermoell/roberta-swedish-scandi}} of ROBERTA \citep{liu2019roberta}
\end{itemize}
The models aimed at generative tasks were more scarce and included:
\begin{itemize}
	\item \citet{ekgren2022lessons} demonstrated early attempts at a Swedish version of GPT-3 \citep{brown2020language}, called GPT-SW3.
	\item a Swedish version\footnote{\url{https://huggingface.co/birgermoell/swedish-gpt}} of GPT-2 \citep{radford2019language};
\end{itemize}  

There have also been attempts at training a Swedish model\footnote{\url{https://huggingface.co/birgermoell/t5-base-swedish}} based on T5 \citep{raffel2020exploring}, which have been shown to perform well for both discriminative and generative tasks for English.

In addition, there have also been attempts to use and evaluate Swedish discriminative models for generative tasks \citep{kalpakchi-boye-2021-bert}, and vice versa \citep{ekgren2022lessons}.

\section{Method}
% TODO: add the word "masked" to "multi-head attention" in Figure 1
\begin{figure*}[!t]
	\centering
	\includegraphics[width=0.7\textwidth]{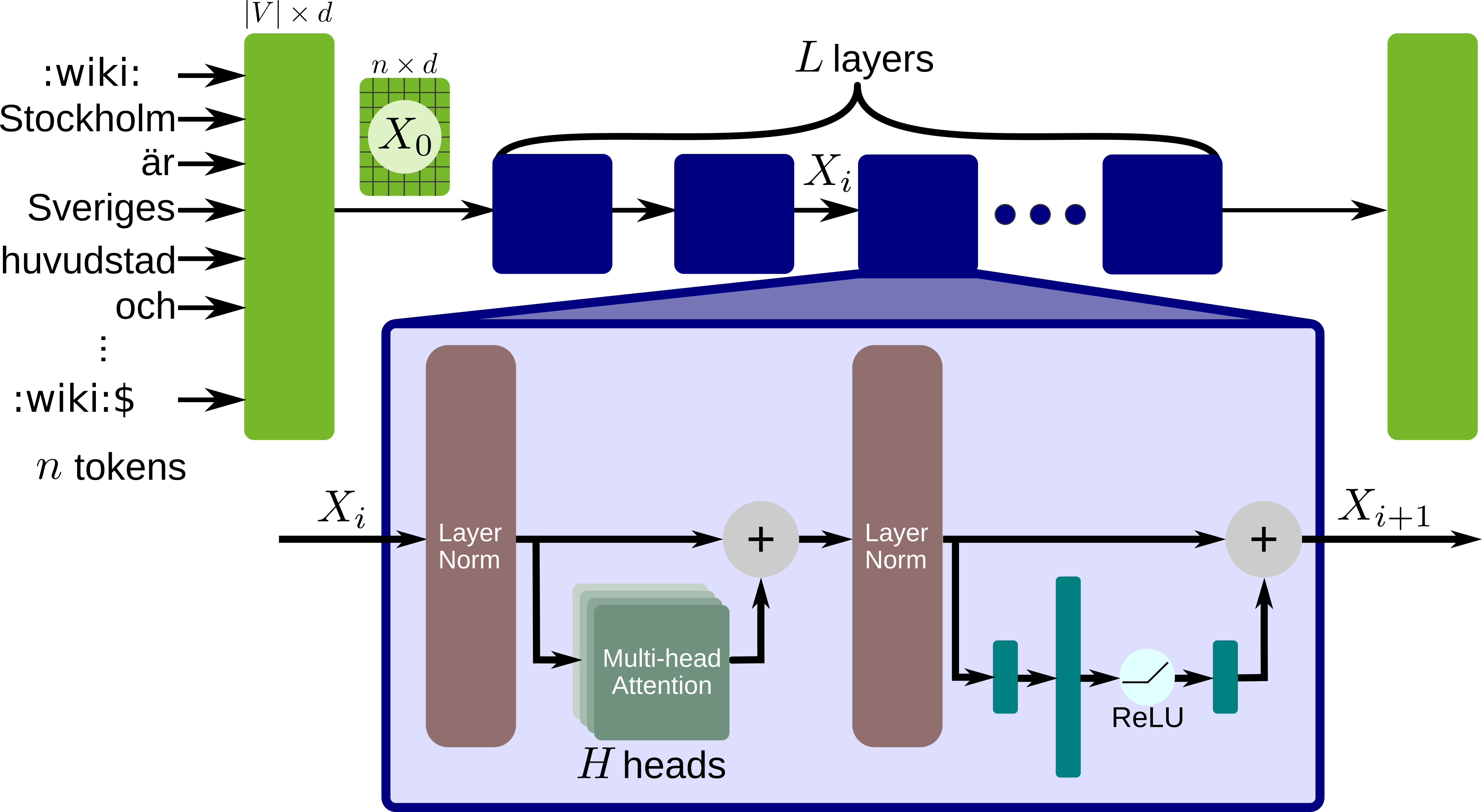}
	\caption{The architecture of SweCTRL-Mini}
	\label{fig:model_arch}
\end{figure*}
The architecture of SweCTRL-Mini, shown in Figure~\ref{fig:model_arch}, is structurally fully based on the CTRL model \citep{keskar2019ctrl}. The idea behind CTRL is to be able to steer the genre of the generated text by using special control codes as 1-token prompts. We will refer to these codes as {\it opening control codes} (OCC). For instance, one could simply provide the OCC \texttt{Wikipedia} (or \texttt{:wiki:} for SweCTRL-Mini) to get a text that looks like a Wikipedia article. One could expand the prompt by providing the beginning of the article, e.g., ``Stockholm is''. Depending on which OCC is put before the words ''Stockholm is'', the genre/style of the generated text should change accordingly.

Compared to the original Transformer model \citep{vaswani2017attention}, it is worth highlighting the following differences in the CTRL architecture:
\begin{itemize}
	\item larger number of heads ($H = 16$ instead of 8);
	\item deeper network ($L = 48$ Transformer blocks instead of 6/12);
	\item tied token embeddings and final output embeddings;
	\item larger vocabulary ($|V| = 256000$ tokens instead of more widespread 50000);
	\item smaller context window ($n = 256$ tokens instead of more widespread 512/1024).
\end{itemize}
For the purpose of this work, we have relied on the implementation of the CTRL model in the Huggingface Transformers library (version 4.20.1) using PyTorch back-end (version 1.12.1).

In order to fulfill our goal of being able to fine-tune on a single GPU, we have \emph{halved} the hidden layer dimensions compared to the original CTRL implementation, more specifically: model dimension $d = 640$, and inner dimension $f = 4096$. These changes make the total number of trainable parameters in SweCTRL-Mini roughly %3 times fewer than 
a third of that of the original English CTRL (something which is also beneficial due to having substantially less training data than for the original model).

Another notable difference is that in addition to the opening control codes (OCC), used by \cite{keskar2019ctrl} to signify different domains (and thus style) of text, we have also added {\it ending control codes} (ECC) to have an indicator of when the model has finished generating text in the given genre. This also allows us to check if the model starts to mix genres, in the case when the sequence starts with an OCC for one genre but ends with an ECC for another genre (see more in Section \ref{sec:gen-hp}). Note that both OCC and ECC were included in the vocabulary (and thus token embeddings) in addition to the already present 256000 tokens.

Similarly to \citet{keskar2019ctrl} we trained the tokenizer using Byte Pair Encoding \citep{bpe1994} on one third of the training data. The tokens for opening, and ending control codes, as well as special tokens for padding and unknown tokens were added on top of the vocabulary of 256000 tokens.

We have trained our model on a node of BerzeLiUs cluster (within the projects Berzelius-2022-161 and Berzelius-2022-169), equipped with 8 NVIDIA A100 GPUs with 40GB of VRAM each. We have used exactly the same hyper-parameters for training as \cite{keskar2019ctrl}, except for setting a global batch size (across GPUs) of 128, instead of the original 1024, due to the constraints in VRAM. The training process took approximately one month and required $181.38$ exaFLOPS (\num{1.8138e+20} FLOPS). In total both training and experimentation towards the final version of our model required $11907.6$ GPU-hours.

\section{Data}
For training we have used 2 data sources: the Swedish part of mC4 \citep{xue2021mt5} and texts from Project Runeberg\footnote{\url{http://runeberg.org/}}. In both cases, however, we have chosen only a subset of data following a number of filtering procedures. These procedures were put in place to ensure data quality to the greatest extent possible given the limited time resources we had at our disposal. Furthermore, we have also applied a number of automatic categorization procedures to classify each text into one of the categories described in Section \ref{sec:cc}. Both kinds of procedures are described in detail in the associated technical note \citep[Section 1]{kalpakchi_dmytro_2023_7868205}. Furthermore, 13-grams (and their substrings) from the training data are searchable through an online interface\footnote{Link is available in the associated GitHub repository.}, as well as the URLs of the web pages included in the part of the mC4 corpus that was used as training data. The found results are accompanied by the category that a text of a specific 13-gram (URL) was assigned to, and whether this category was automatically or manually assigned.

\subsection{Control codes}\label{sec:cc}
The 37 content categories specified in Table \ref{tab:data_by_cat} were used as control codes when training the SweCTRL-Mini model. The distribution of the training data among the categories and their corresponding control codes is shown in Table \ref{tab:data_by_cat}. The categories in \textbf{bold} from this table are later referred to as \emph{major} categories, whereas all the other are referred to as \emph{minor}.

One of the original categories, \emph{blogs/tech} have been completely filtered out by our automatic filtering and categorization procedures (something we didn't notice until after having trained the model). However, the category was also assigned OCC and ECC (and the respective trainable embeddings), hence we keep it both in Table \ref{tab:data_by_cat} and in all further analysis to test the behavior of the model on the ``untrained'' control codes. We will refer to this category as \emph{orphan}.

\begin{table*}
	\centering
	\begin{tabular}{lrrr}
		\toprule
		Category            & Documents &     Tokens &  Characters \\
		\midrule
		\textbf{news}       & 1 629 526 &  635 179 726 &  4 060 209 799 \\
		\textbf{wiki}       &   412 421 &  151 181 708 &  1 095 556 624 \\
		news/sport          &   358 016 &  149 461 664 &    903 154 064 \\
		\textbf{forum}      &   316 664 &  212 420 326 &  1 244 181 506 \\
		\textbf{blogs}      &   297 258 &  205 834 053 &  1 179 533 133 \\
		news/pressrelease   &   277 017 &   88 914 953 &    621 764 444 \\
		\textbf{ads}        &   260 959 &   84 948 629 &    587 084 926 \\
		news/opinion        &   221 010 &  113 723 324 &    730 484 523 \\
		news/culture        &   150 241 &   66 817 491 &    419 183 009 \\
		\textbf{admin}      &   136 495 &  169 036 110 &  1 185 927 529 \\
		news/economy        &    76 421 &   26 637 789 &    174 005 712 \\
		\textbf{debate}     &    67 831 &   68 290 441 &    468 393 340 \\
		info/medical        &    42 952 &   18 829 692 &    122 089 820 \\
		\textbf{info}       &    34 035 &   12 650 622 &     82 901 241 \\
		news/tech           &    30 004 &    8 083 200 &     49 342 576 \\
		\textbf{review}     &    24 017 &   11 614 089 &     71 752 874 \\
		info/travel         &    21 528 &    7 713 750 &     46 624 552 \\
		forum/law           &    20 982 &   13 007 375 &     79 398 374 \\
		news/lifestyle      &    20 978 &   13 022 322 &     78 661 676 \\
		blogs/sport         &    13 134 &    9 613 961 &     58 007 976 \\
		info/lifestyle      &    13 056 &    5 780 355 &     35 801 351 \\
		news/sustainability &    12 975 &    3 951 222 &     26 320 596 \\
		forum/sport         &    12 649 &    9 747 421 &     56 947 940 \\
		forum/tech          &    12 286 &    4 899 979 &     30 984 736 \\
		news/travel         &    10 118 &    6 555 937 &     41 146 765 \\
		info/business       &     8 793 &    4 649 150 &     28 408 960 \\
		news/politics       &     7 683 &    1 870 196 &     12 544 739 \\
		news/science        &     7 295 &    2 849 480 &     17 981 928 \\
		news/food           &     5 893 &    2 415 800 &     14 831 815 \\
		forum/travel        &     3 844 &    1 632 462 &     10 272 658 \\
		news/fashion        &     3 278 &    1 665 669 &      9 610 223 \\
		news/weather        &       841 &      477 327 &      2 928 822 \\
		blogs/economy       &       672 &      343 747 &      2 214 424 \\
		forum/economy       &       340 &      329 584 &      2 051 113 \\
		\textbf{\textit{literature}} &       297 & 	  1 992 736 &     12 274 721 \\
		\textbf{simple}              &           25 &        9 618 &         56 865 \\
		blogs/tech                   &            0 &            0 &              0 \\
		\bottomrule
	\end{tabular}
	\caption{\label{tab:data_by_cat} Distribution of training data by category/control code. The categories in \textbf{bold} are referred to as \emph{major} categories, whereas all the other are referred to as \emph{minor}. The category in \textit{italics}, literature, was the only one trained on data from Project Runeberg. All other categories were trained on mC4.}
\end{table*}

\section{Model selection}
Selecting a model includes two decisions: choosing a version of the model from the training (a so-called {\it checkpoint}) (Section \ref{sec:perp}), and deciding on the default hyper-parameters for the generation process (Section \ref{sec:gen-hp}).

\subsection{Estimating dataset overlap}
A cornerstone of evaluation is that only data unseen during training should be used as test data. If the training corpora include a lot of text, there is a higher risk that substantial portions of texts overlap between the training and evaluation corpora. 

To get an insight into this issue, we attempted to estimate the overlap between the texts in each evaluation dataset $D$ and texts from the training data $T$. In order to do this, we simply counted the frequency $O_{D,T,F}^k$ of $k$-grams being present both in $D$ and $T$ at a certain frequency threshold $F$, as follows:
\begin{equation*}
	O_{D,T,F}^k = \frac{|ng_k: ng_k \in D \land tf_{T, ng_k} \ge F|}{|ng_k: ng_k \in D|} \cdot 100\%
\end{equation*}
For the brevity of notation, we will omit $D$ and $T$ and denote $O_{D,T,F}^k$ as $O_F^k$.

In order to choose a checkpoint, we have calculated what perplexity the model assigns to texts from 4 datasets: the training sets of UD-LinES \citep{ahrenberg2007lines}, UD-Talbanken \citep{nivre2007bootstrapping}, SweQUAD-MC \citep{kalpakchi-boye-2021-bert}, and the validation set of the Swedish part of mC4 (subject to the same filtering as the training set and thus denoted as mC4*). Note that we have used the training sets of the 3 datasets effectively as validation sets for SweCTRL-Mini, simply because the training sets include more text. Also note that \emph{none} of these datasets were in our training data explicitly; however, the texts could still be similar, which could result in an unfair advantage. In order to check that, we have reported the sizes of the overlaps between each of the 4 aforementioned datasets and the training set of SweCTRL-Mini in Table~\ref{tab:overlap_perplexity}, for 7-grams and 13-grams respectively.

\subsection{Perplexity}
\label{sec:perp}
\begin{figure}[!t]
	\centering
	\includegraphics[width=0.5\linewidth]{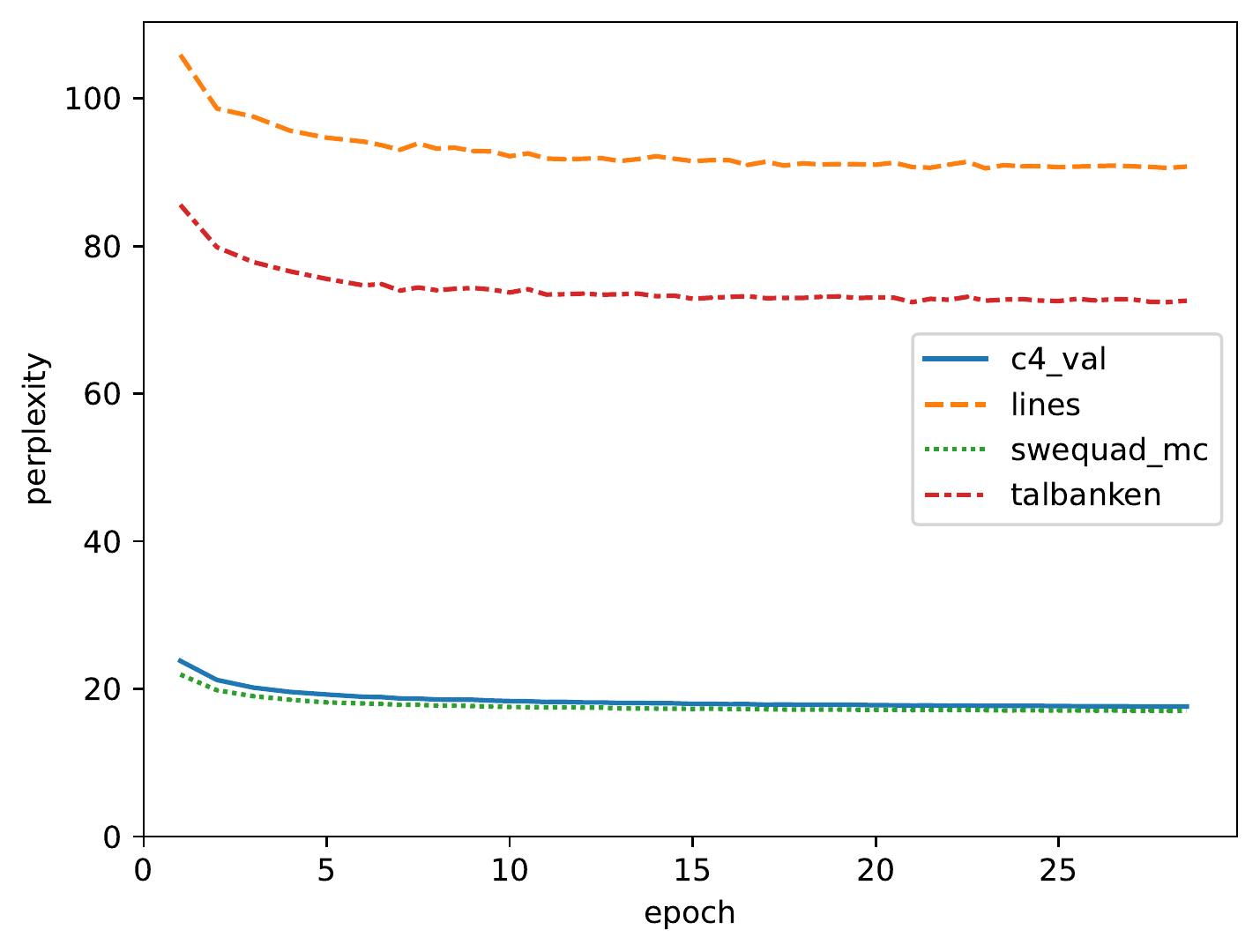}
	\caption{Perplexity $PP_w$ of the 52 different SweCTRL-Mini checkpoints}
	\label{fig:ppl}
\end{figure}

\begin{table*}[!t]
	\centering
	\begin{tabular}{lcccccccc}
		\toprule
		\multirow{2}{*}{\textbf{Dataset}} & \multicolumn{4}{c}{7-grams} & \multicolumn{4}{c}{13-grams} \\
		& \textbf{$N_{<7}$} & \textbf{$O_1^7$} & \textbf{$O_{10}^7$} & \textbf{$O_{100}^7$} &  \textbf{$N_{<13}$} & \textbf{$O_1^{13}$} & \textbf{$O_{10}^{13}$} & \textbf{$O_{100}^{13}$} \\
		\midrule
		mC4* (v) & 0.4\% & 15.86\% & 2.55\% & 0.99\% & 0.8\% & 9.93\% & 0.99\% & 0.39\% \\
		UD-T (tr) & 18.8\% & 1.35\% & 0.23\% & 0.03\% & 50.6\% & 0.03\% & 0\% & 0\% \\
		UD-L (tr) & 17.3\% & 1.16\% & 0.15\% & 0.02\% & 46.1\% & 0.09\% & 0\% & 0\% \\
		S-MC (tr) & 0\% & 6.82\% & 5.51\% & 0.07\% & 0\% & 3.18\% & 0.04\% & 0\% \\
		\bottomrule
	\end{tabular}
	\caption{Overlap in 7-grams and 13-grams between the training set of SweCTRL-Mini and the specified split of each dataset. $N_{<7}$ ($N_{<13}$) corresponds to the proportion of texts shorter than 7 (13) tokens, all $O^7_*$ ($O^{13}_*$) metrics are proportions of the texts longer than 7 (13) tokens, out of $|D| - N_{<7}$ ($|D| - N_{<13}$)}\label{tab:overlap_perplexity}
\end{table*}
Perplexity was calculated using the sliding window approach\footnote{\url{https://huggingface.co/docs/transformers/perplexity}}:
\begin{equation}\label{eq:pp}
	\widetilde{PP_w} = \exp\bigg\{-\frac{1}{T}\sum_{i=0}^{T}\ln p(x_i|x_{i-w:i-1})\bigg\}
\end{equation}
The results are reported in Figure \ref{fig:ppl}.
Note that when $\widetilde{PP_w}$ on the dataset $D$ decreases, so does the cross-entropy loss on $D$. Hence, perplexity on the validation set is effectively a proxy to the validation loss and can therefore be used to track overfitting. For all 4 datasets, there is a weak decreasing trend in perplexities, suggesting that the model has not overfitted to the training data. After about the 15th epoch, the perplexity has more or less plateaued with very slight variations in values. Such small differences can be dataset-specific, and a slightly lower perplexity (a difference in the order of 0.01) does not necessarily entail a better checkpoint. Because of this observation, we chose to use the last available checkpoint for all further experiments.

\subsection{Search for generation hyper-parameters}
\label{sec:gen-hp}
SweCTRL-Mini, as the original CTRL, is an autoregressive model using left-to-right generation, meaning that text generation happens one token at a time. The generated words are decided based on the softmax distribution for the next token provided by the model. The established way of generating a new token is to sample from this distribution, notably to avoid generating text in a loop, a phenomenon we will refer to as \emph{sampling loops}. Sampling can be performed in a multitude of ways, where the resulting distribution is tweaked via \emph{generation hyper-parameters} to increase/decrease certain probabilities or to limit the number of tokens considered for selection. In this paper, we attempted to perform a systematic search through some of these hyper-parameters, specifically temperature $T \in (0, 1]$,  repetition penalty $r \in [1, 2]$ (\citet{keskar2019ctrl}), and the nucleus threshold $p \in [0.7, 1]$ (\citet{holtzman2019curious}).

To investigate which hyper-parameter settings work best for SweCTRL-Mini, we have employed a grid search over generation hyper-parameters in \emph{pairs}: $(p, r)$, and $(T, r)$. The rationale behind such hyper-parameter pairing is that $p$ and $T$ attempt to achieve the same goal: denying sampling of highly improbable tokens (with a varying definition of ``highly improbable''). On the other hand, $r$ has a different goal: prevent sampling loops, and, as a by-product, increase lexical diversity. We have evaluated the results of the grid search using the following automatic metrics:
\begin{enumerate}
	\item \textbf{Presence of ECC}. The idea with ECC is that SweCTRL-Mini should learn how to start and end texts in each genre appropriately. Hence, when the OCC and ECC do not match, this might potentially indicate a topic/genre shift, which is undesirable.
	\item \textbf{Size of sampling loops}. We have calculated the number of contiguously repeating phrases (up to the length of 5 tokens). Evidently, no such phrases are desirable at all, except, if for the case when \emph{all} repeated tokens are numerals (such as ``22'' being tokenized as double ``2''), which was excluded from the analysis.
	\item \textbf{Number of generated tokens}. There are no clear-cut rules for this metric, but our goal was to get as large a variance as possible in this category, to be able to generate texts of varying lengths.
	\item \textbf{BLEU-4}, introduced by \cite{papineni2002bleu}. Here, we actually aim for a BLEU-4 score that is as low as possible, in order to maximize the lexical diversity of the generated texts. 
\end{enumerate}

A detailed account of the methodology and results of the grid search is provided in the technical note \citep[Section 3]{kalpakchi_dmytro_2023_7868205}. To summarize the results, we observed  the following trends:
\begin{itemize}
	\item the larger the repetition penalty $r$, the larger the variance in the lengths of generated texts, up until some value $r'$ after which the variance remains more or less constant;
	\item if there is no repetition penalty ($r = 1.0$), the model tends to produce a substantial number of repeated phrases (up to length of 5);
	\item fully duplicated texts are extremely rare (except when $T = 0$, which corresponds to deterministic argmax-generation, where it is fully expected);
	\item most texts either do not reach any control code, or reach the correct ending control code, except for categories ``Literature'' and ``Simple'', indicating that the model is most probably incapable of generating texts without topic/style shifts in these categories.
\end{itemize}

Additionally, we observed that no combination of sampling hyper-parameters dominates the others according to our automatic evaluation metrics. The combinations that seem to perform well across most categories are \texttt{M1}: $r = 1.6, p = 0.8$, and \texttt{M2}: $r = 1.4, p = 0.9$. In addition, we also deem \texttt{M3}: $r = 1.0, p = 0.9$ interesting, although \texttt{M3} does not perform favorably on our automatic metrics. However, our decisions so far have been based on automatic metrics, and one of the ways to validate them is to conduct human evaluation on both well- and not-so-well-performing combinations of generation hyper-parameters. 

\begin{figure*}[!ht]
	\begin{subfigure}{\textwidth}
		\centering
		\includegraphics[width=\textwidth]{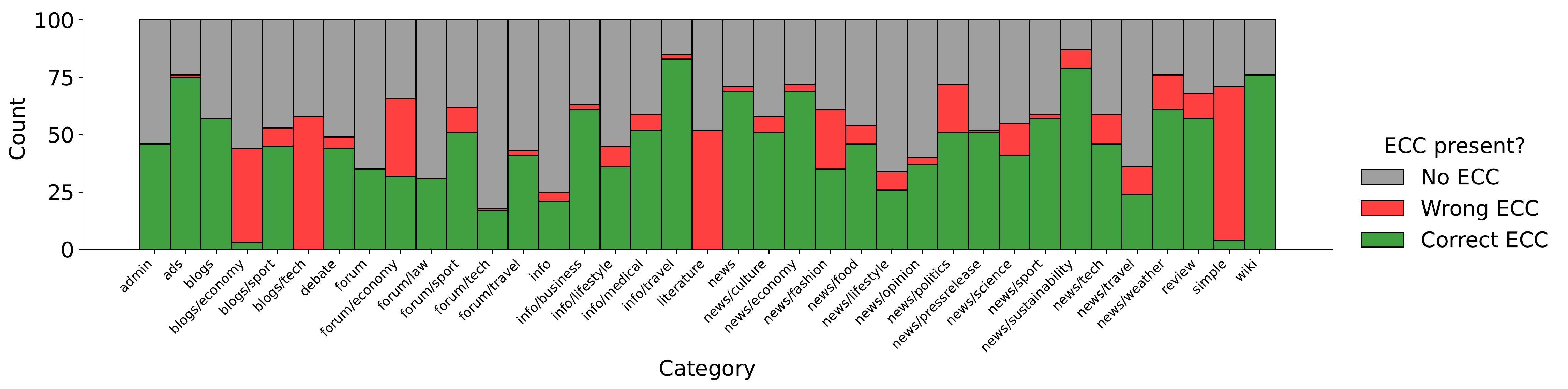}
		\vspace{-0.5em}
		\caption{the endings of the generated texts}
		\vspace{0.6em}
		\label{fig:ecc_all_16_08}
	\end{subfigure}
	
	\begin{subfigure}{\textwidth}
		\centering
		\includegraphics[width=0.99\textwidth]{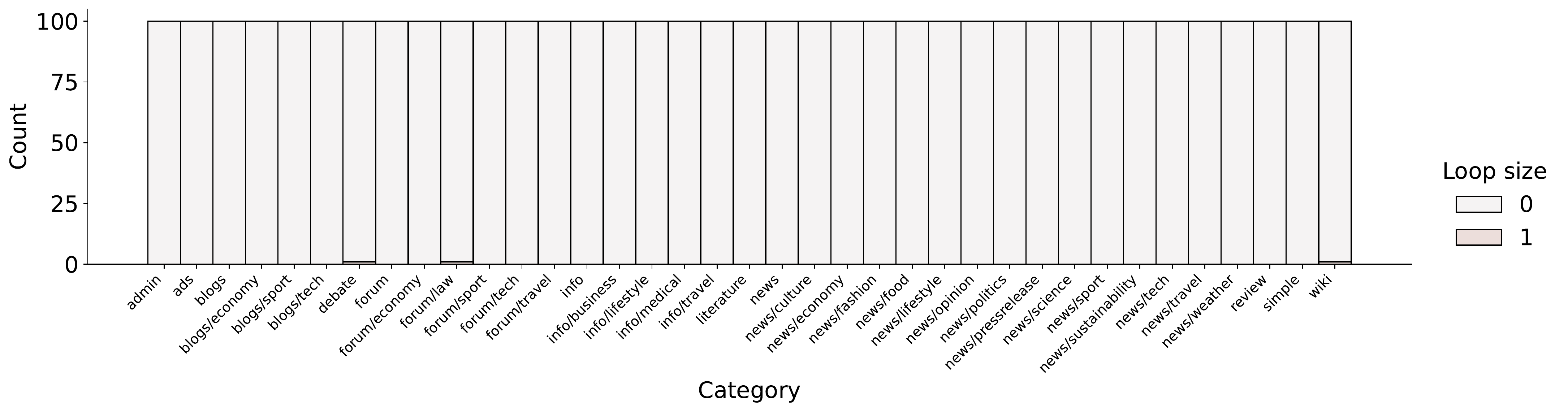}
		\vspace{-0.5em}
		\caption{the sampling loop size (up to 5 tokens)}
		\vspace{0.6em}
		\label{fig:repeats_all_16_08}
	\end{subfigure}
	
	\begin{subfigure}{\textwidth}
		\centering
		\includegraphics[width=0.95\textwidth]{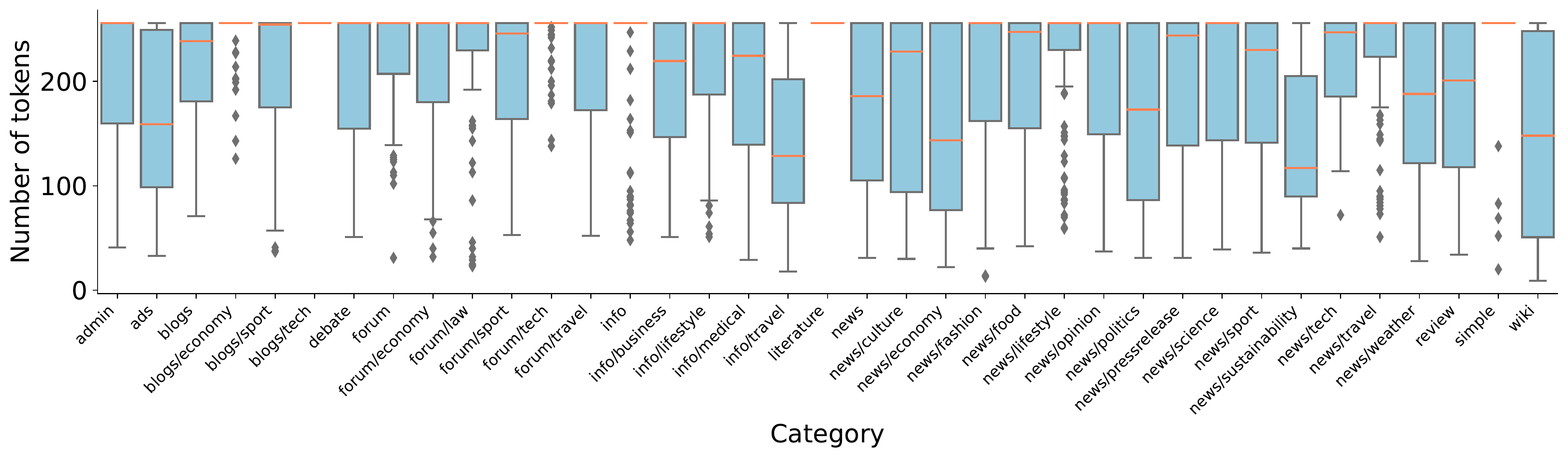}
		\vspace{-0.5em}
		\caption{the number of generated tokens}
		\vspace{0.6em}
		\label{fig:tokens_all_16_08}
	\end{subfigure}	
	
	\begin{subfigure}{\textwidth}
		\centering
		\includegraphics[width=0.95\textwidth]{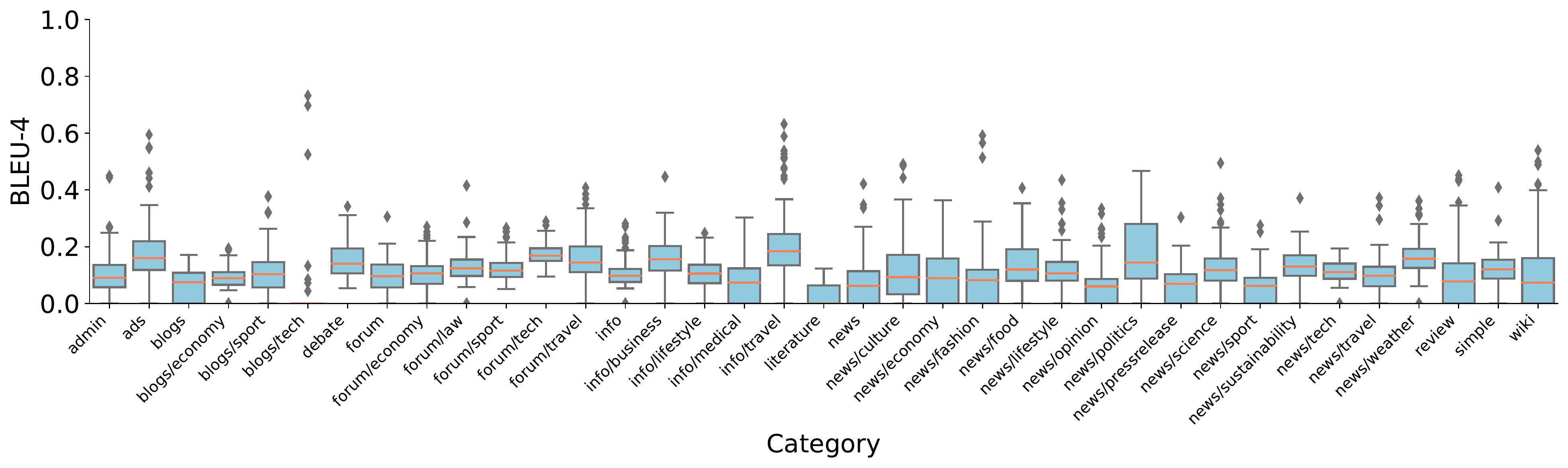}
		\vspace{-0.5em}
		\caption{BLEU-4 scores for sampled texts within each category (the lower, the better)}
		\vspace{0.6em}
		\label{fig:bleu_all_16_08}
	\end{subfigure}
	\vspace{-1em}
	\caption{Automatic metrics for the generated texts using \texttt{M1} ($r = 1.6$, and $p = 0.8$)}\label{fig:all_16_08}
\end{figure*}

\begin{figure*}[!ht]
	\begin{subfigure}{\textwidth}
		\centering
		\includegraphics[width=\textwidth]{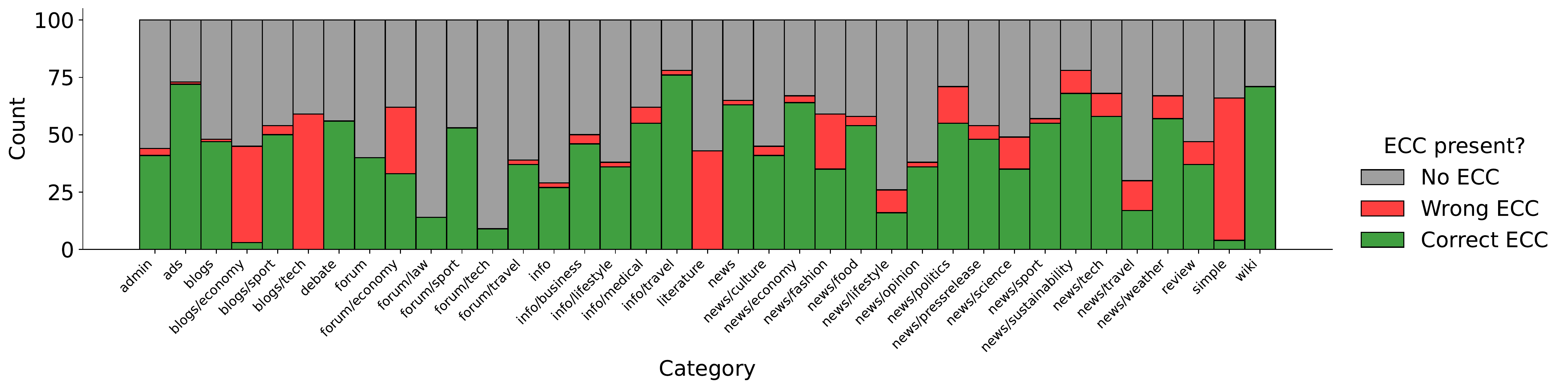}
		\vspace{-0.5em}
		\caption{the endings of the generated texts}
		\vspace{0.6em}
		\label{fig:ecc_all_14_09}
	\end{subfigure}
	
	\begin{subfigure}{\textwidth}
		\centering
		\includegraphics[width=0.99\textwidth]{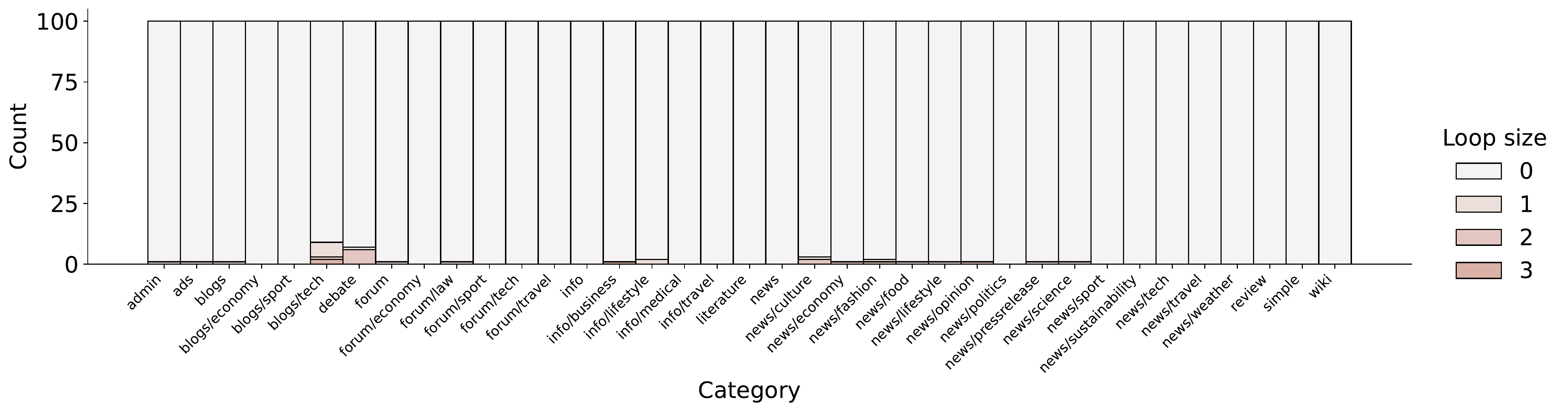}
		\vspace{-0.5em}
		\caption{the sampling loop size (up to 5 tokens)}
		\vspace{0.6em}
		\label{fig:repeats_all_14_09}
	\end{subfigure}
	
	\begin{subfigure}{\textwidth}
		\centering
		\includegraphics[width=0.95\textwidth]{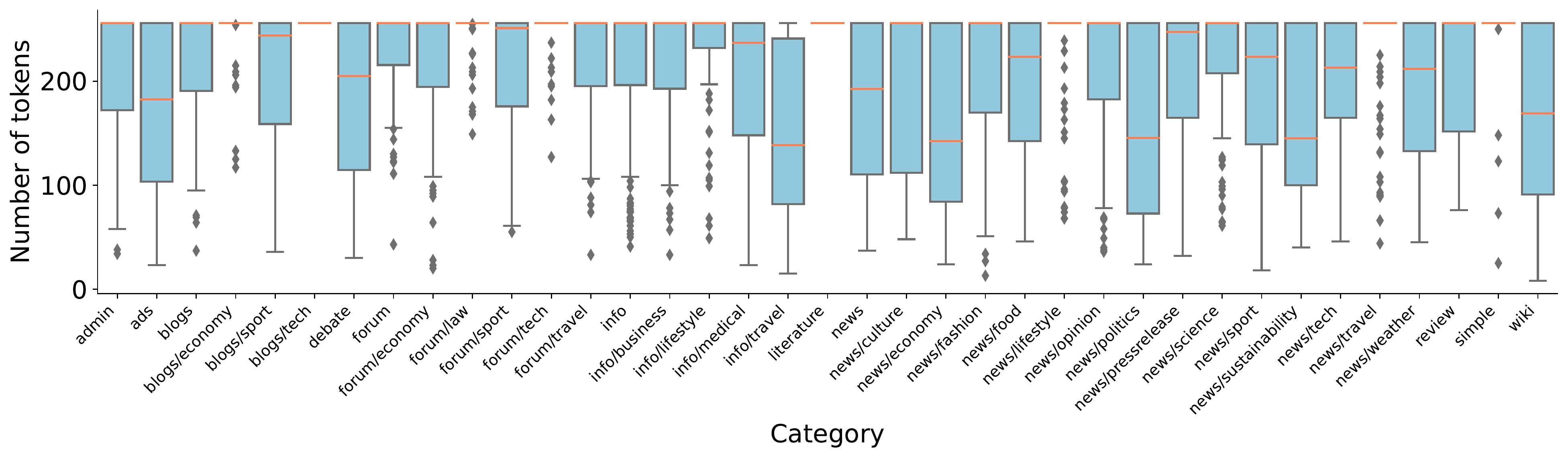}
		\vspace{-0.5em}
		\caption{the number of generated tokens}
		\vspace{0.6em}
		\label{fig:tokens_all_14_09}
	\end{subfigure}	
	
	\begin{subfigure}{\textwidth}
		\centering
		\includegraphics[width=0.95\textwidth]{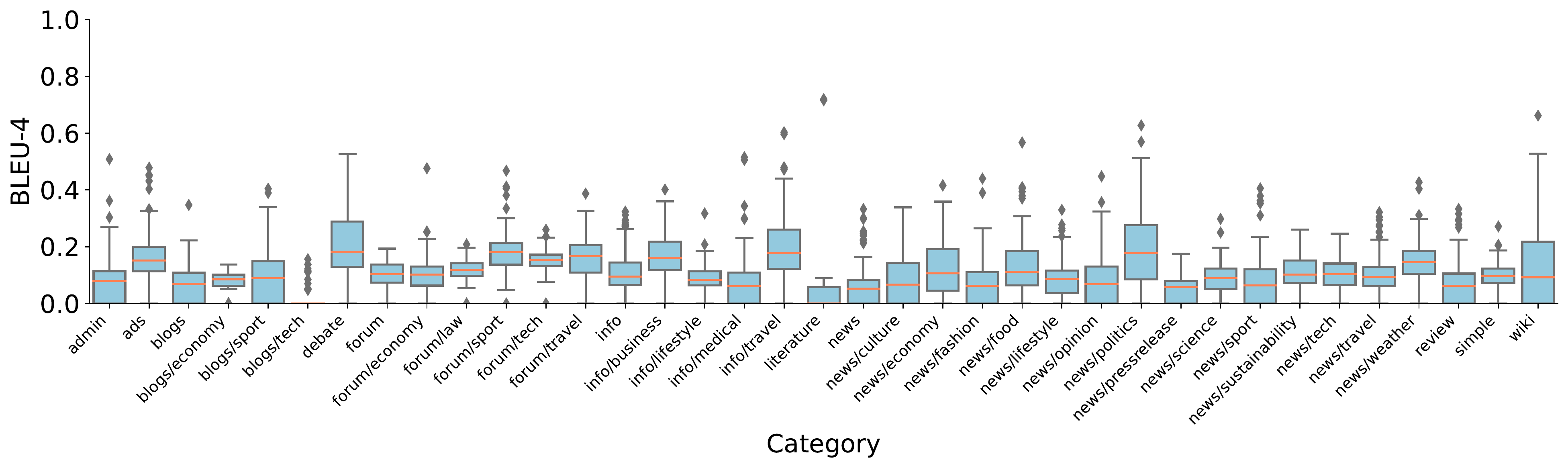}
		\vspace{-0.5em}
		\caption{BLEU-4 scores for sampled texts within each category (the lower, the better)}
		\vspace{0.6em}
		\label{fig:bleu_all_14_09}
	\end{subfigure}
	\vspace{-1em}
	\caption{Automatic metrics for the generated texts using \texttt{M2} ($r = 1.4$, and $p = 0.9$)}\label{fig:all_14_09}
\end{figure*}

\begin{figure*}[!ht]
	\begin{subfigure}{\textwidth}
		\centering
		\includegraphics[width=\textwidth]{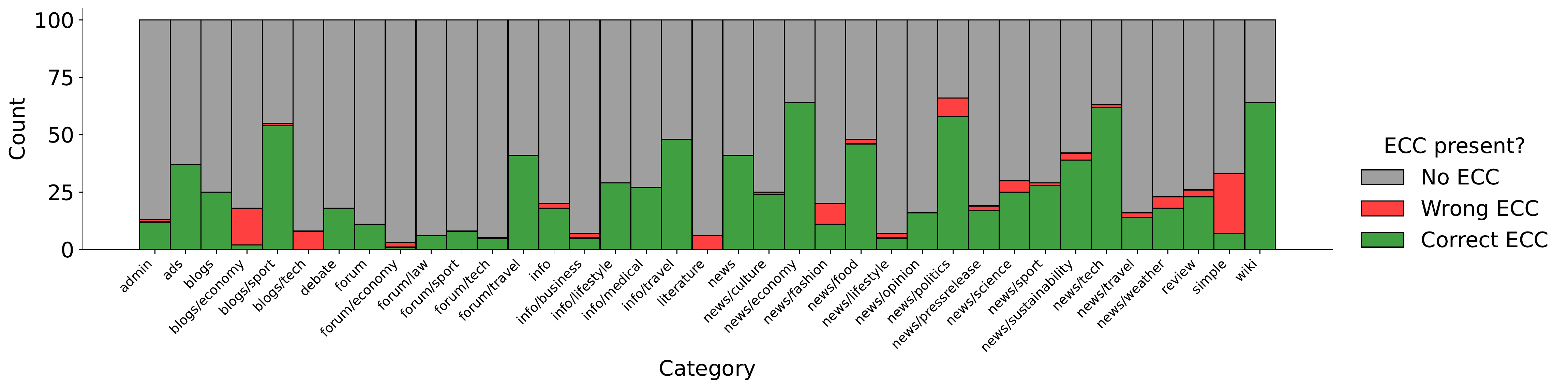}
		\vspace{-0.5em}
		\caption{the endings of the generated texts}
		\vspace{0.6em}
		\label{fig:ecc_all_10_09}
	\end{subfigure}
	
	\begin{subfigure}{\textwidth}
		\centering
		\includegraphics[width=0.99\textwidth]{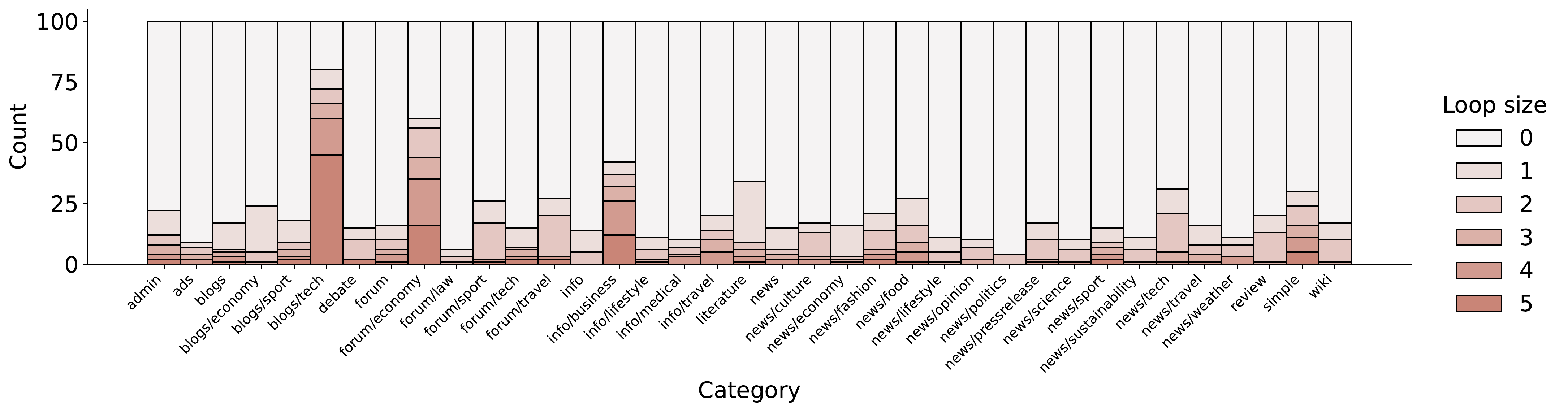}
		\vspace{-0.5em}
		\caption{the sampling loop size (up to 5 tokens)}
		\vspace{0.6em}
		\label{fig:repeats_all_10_09}
	\end{subfigure}
	
	\begin{subfigure}{\textwidth}
		\centering
		\includegraphics[width=0.95\textwidth]{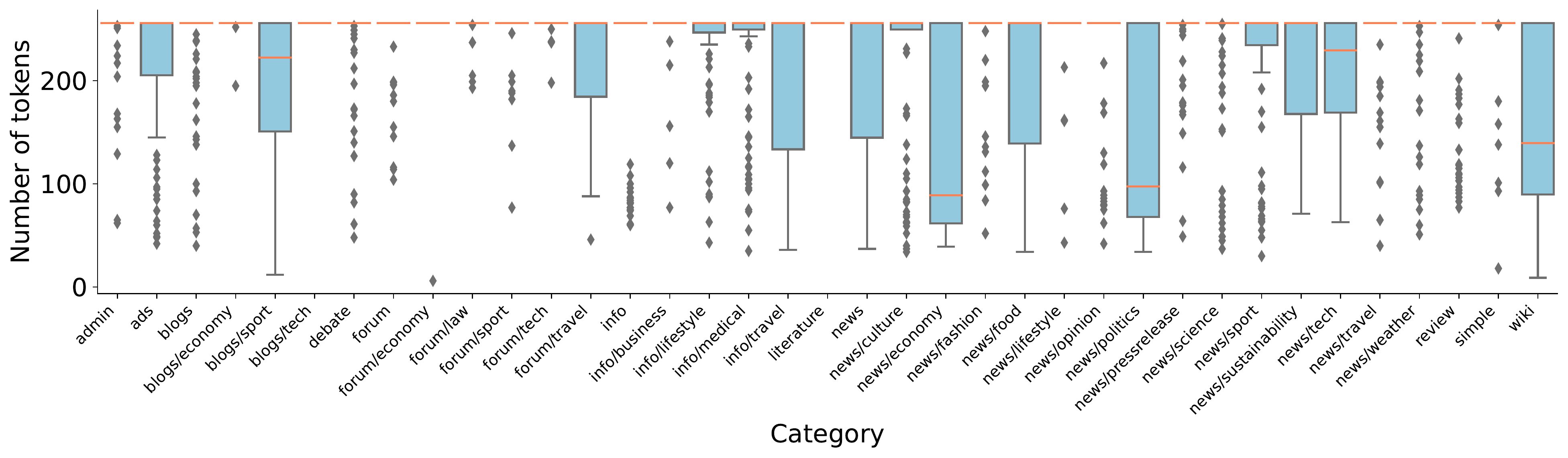}
		\vspace{-0.5em}
		\caption{the number of generated tokens}
		\vspace{0.6em}
		\label{fig:tokens_all_10_09}
	\end{subfigure}	
	
	\begin{subfigure}{\textwidth}
		\centering
		\includegraphics[width=0.95\textwidth]{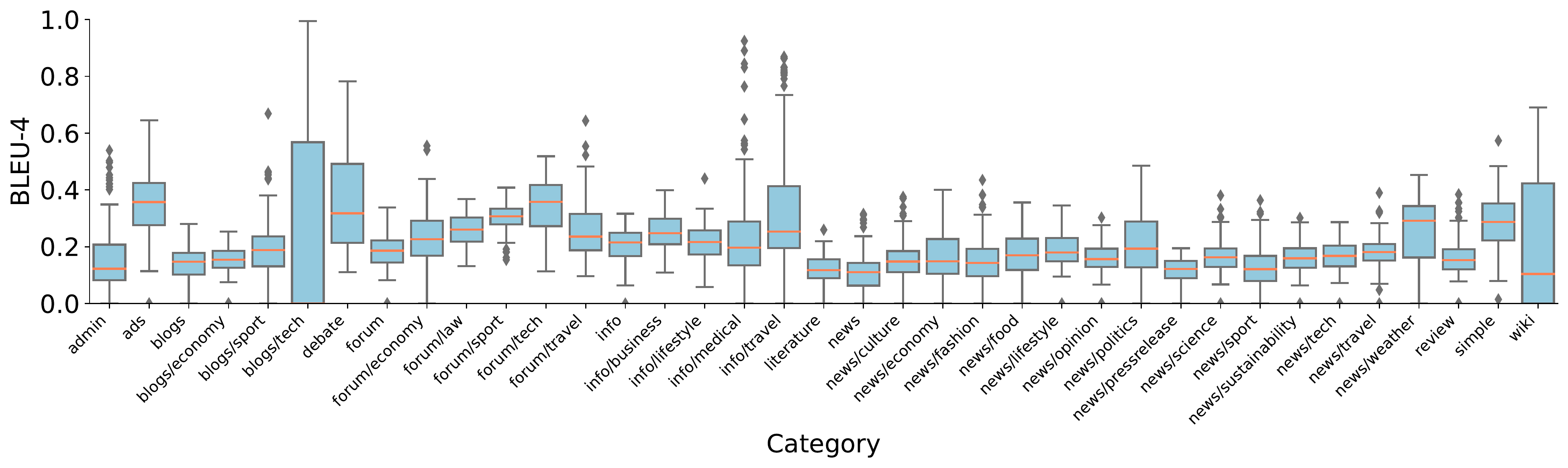}
		\vspace{-0.5em}
		\caption{BLEU-4 scores for sampled texts within each category (the lower, the better)}
		\vspace{0.6em}
		\label{fig:bleu_all_10_09}
	\end{subfigure}
	\vspace{-1em}
	\caption{Automatic metrics for the generated texts using \texttt{M3} ($r = 1.0$, and $p = 0.9$)}\label{fig:all_10_09}
\end{figure*}

\begin{figure*}
	\begin{subfigure}{\textwidth}
		\centering
		\includegraphics[width=0.95\textwidth]{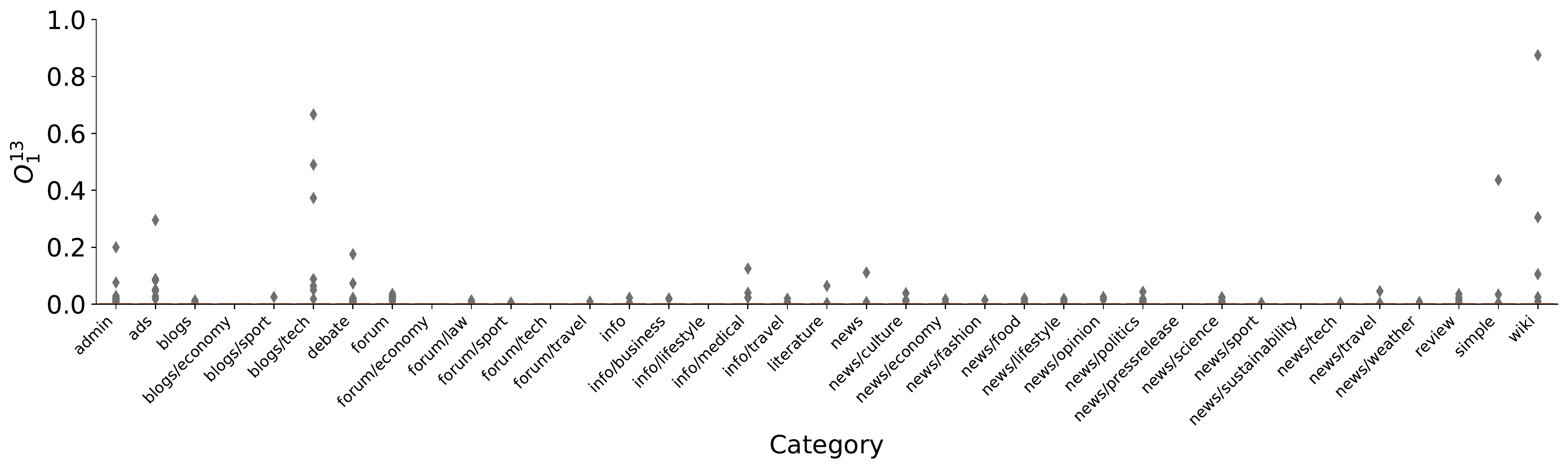}
		\vspace{-0.5em}
		\caption{\texttt{M1} ($r = 1.6$, and $p = 0.8$)}
		\vspace{0.6em}
		\label{fig:overlap_all_16_08}
	\end{subfigure}
	
	\begin{subfigure}{\textwidth}
		\centering
		\includegraphics[width=0.95\textwidth]{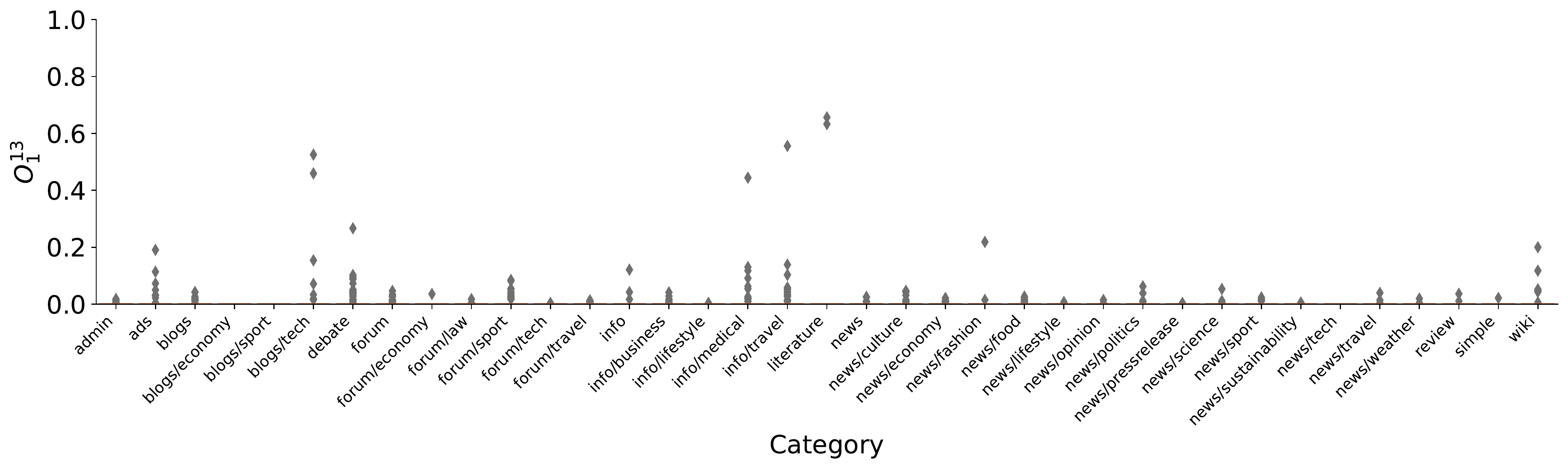}
		\vspace{-0.5em}
		\caption{\texttt{M2} ($r = 1.4$, and $p = 0.9$)}
		\vspace{0.6em}
		\label{fig:overlap_all_14_09}
	\end{subfigure}	
	
	\begin{subfigure}{\textwidth}
		\centering
		\includegraphics[width=0.95\textwidth]{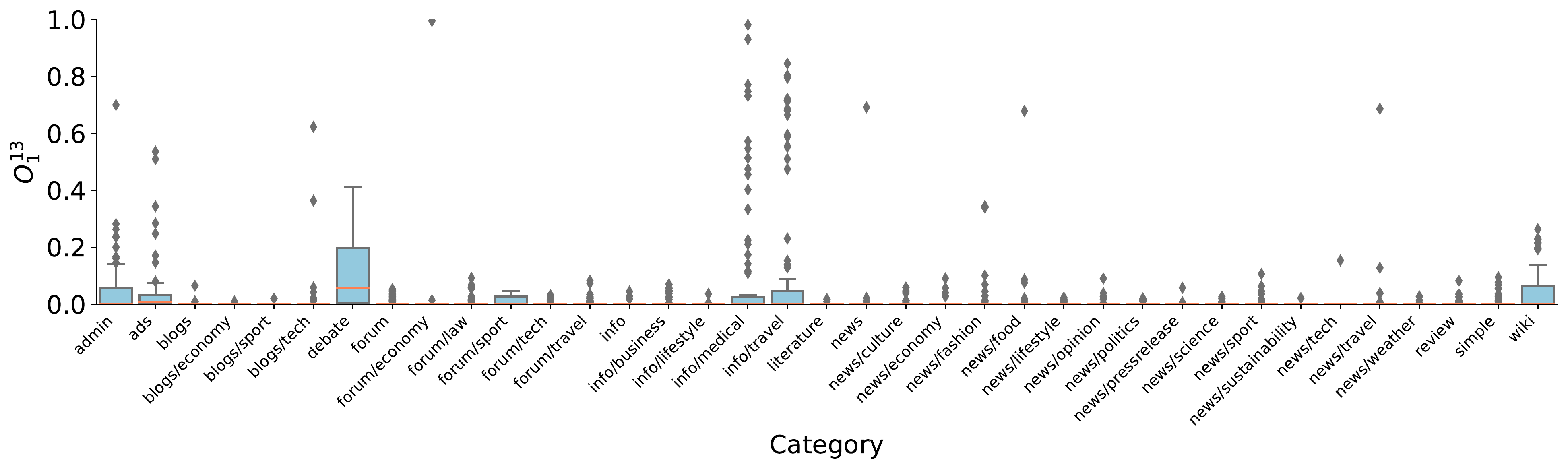} 
		\vspace{-0.5em}
		\caption{\texttt{M3} ($r = 1.0$, and $p = 0.9$)}
		\vspace{0.6em}
		\label{fig:overlap_all_10_09}
	\end{subfigure}
	\caption{Box plots (with 1.5 IQR whiskers) showing 13-gram overlap $O_1^{13}$ between the produced texts for each category and SweCTRL's training data (the lower, the better)}\label{fig:overlap_all}
\end{figure*}

Before proceeding with human evaluation for these combinations (see Section \ref{sec:human-eval}), we conducted a further automatic evaluation for the minor categories to inform our category selection for human evaluation. We have conducted the same kind of analysis as for major categories in subplots (a) - (d) in Figures \ref{fig:all_16_08}, \ref{fig:all_14_09}, and \ref{fig:all_10_09} for \texttt{M1}, \texttt{M2}, and \texttt{M3}, respectively. Additionally, we have looked at the 13-gram overlap between the generated texts and the training data of SweCTRL-Mini in Figure \ref{fig:overlap_all}.

\begin{figure*}[!ht]
	\centering
	\includegraphics[width=0.75\textwidth]{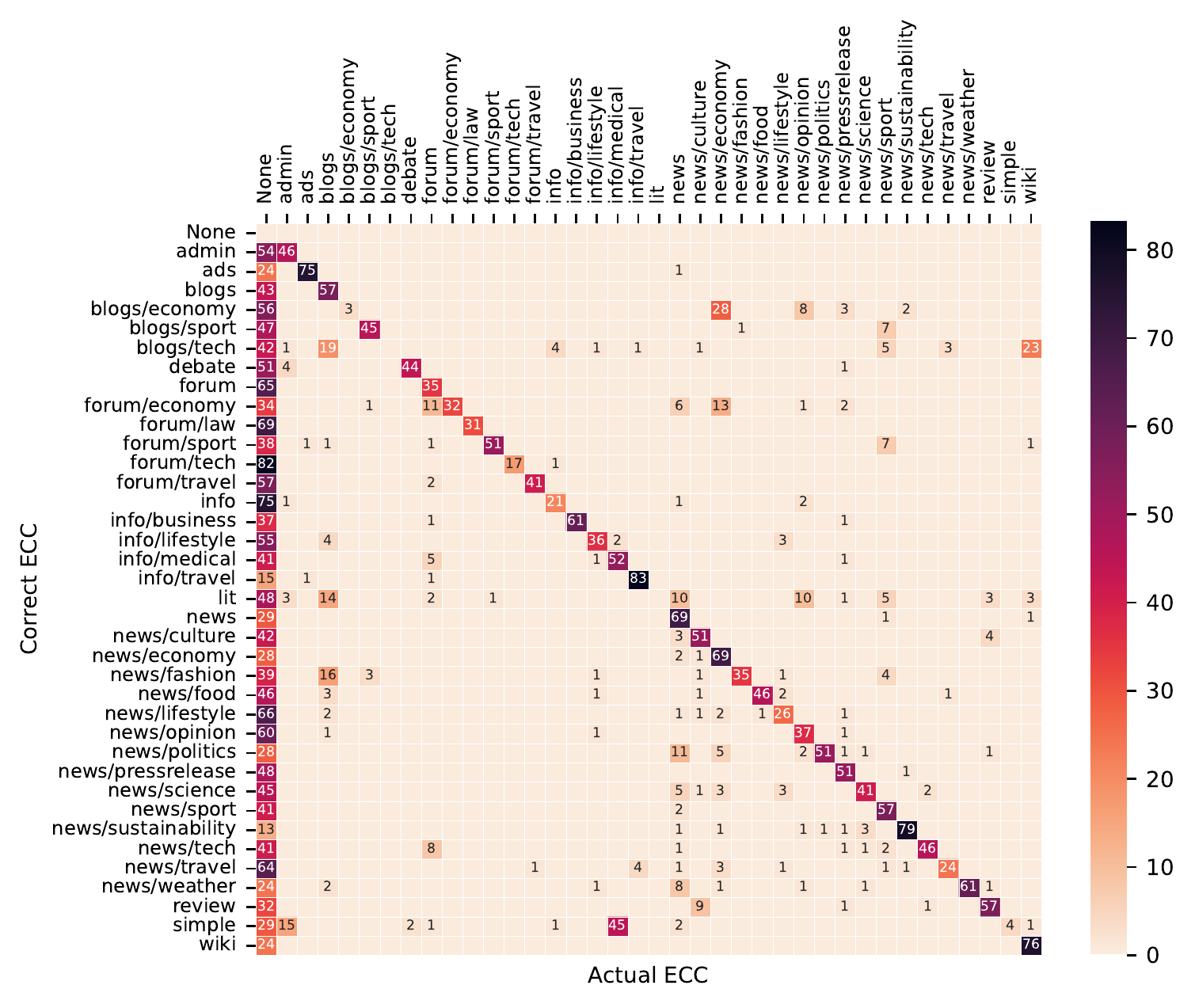}
	\caption{The confusion matrix for ECC of the generated texts (\texttt{M1})}
	\label{fig:conf_matrix_16_08}
\end{figure*}

\begin{figure*}[!ht]
	\centering
	\includegraphics[width=0.75\textwidth]{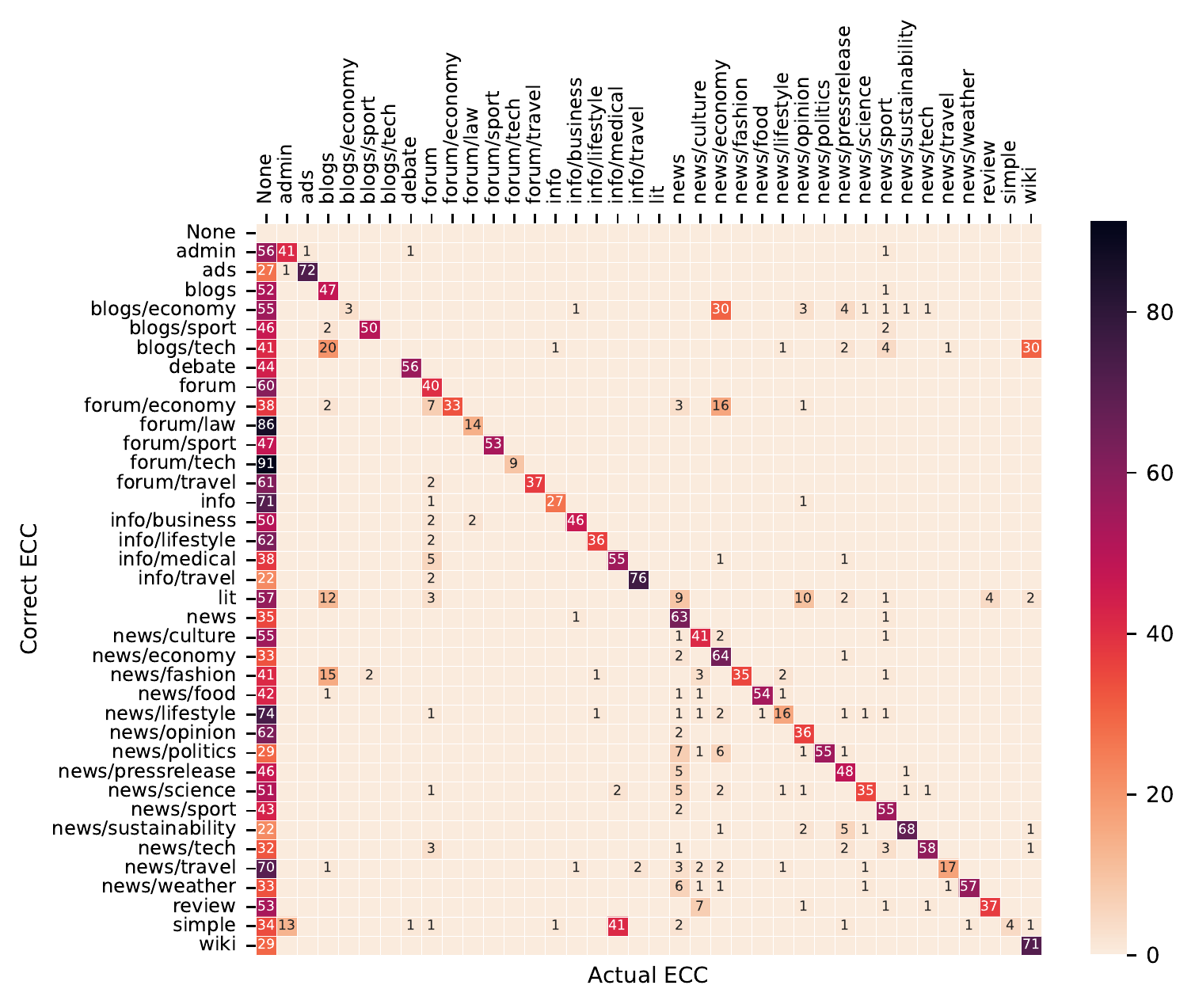}
	\caption{The confusion matrix for ECC of the generated texts (\texttt{M2})}
	\label{fig:conf_matrix_14_09}
\end{figure*}

\begin{figure*}[!ht]
	\centering
	\includegraphics[width=0.75\textwidth]{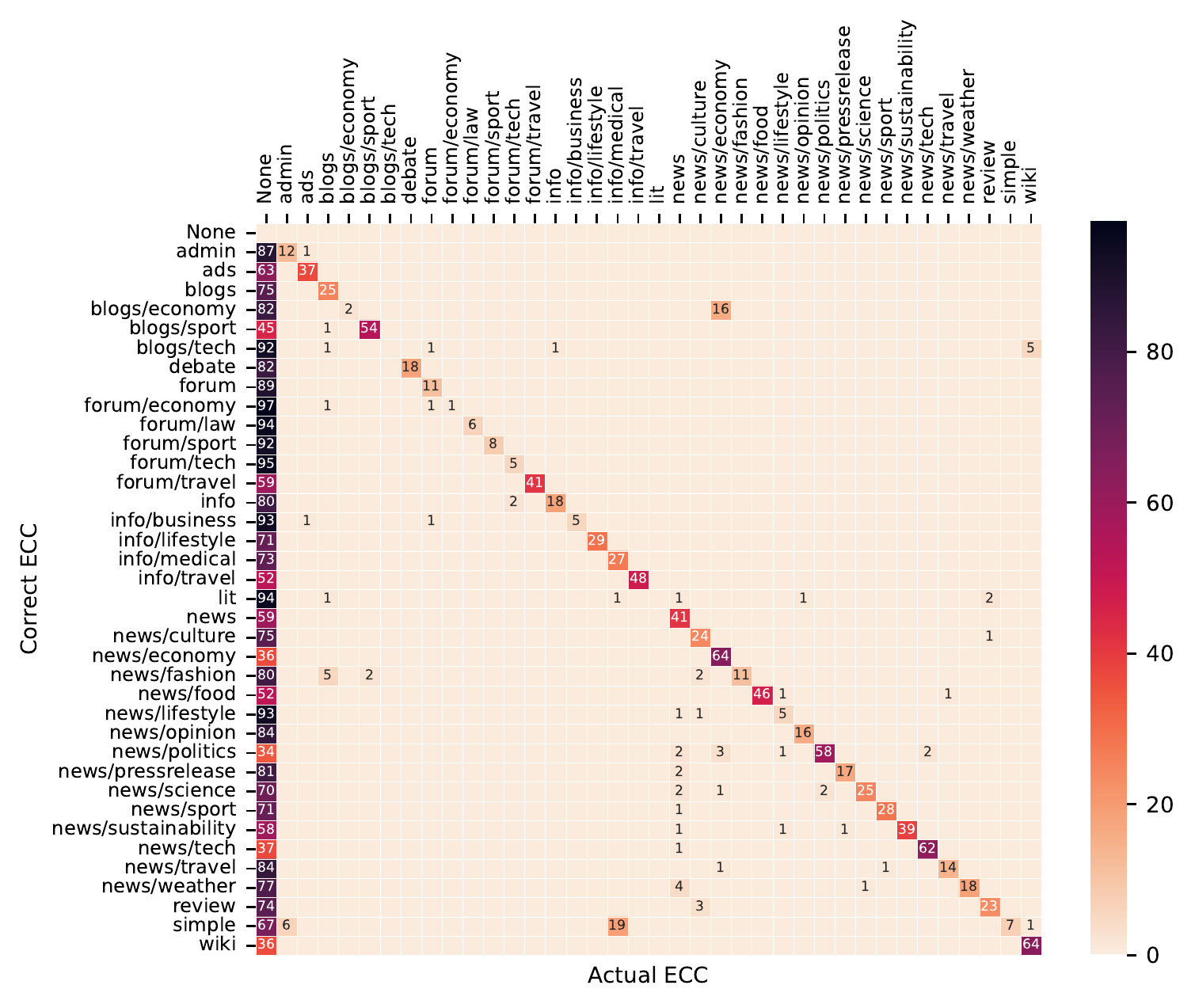}
	\caption{The confusion matrix for ECC of the generated texts (\texttt{M3})}
	\label{fig:conf_matrix_10_09}
\end{figure*}

We first look at the orphan category, \emph{blogs/tech}, and immediately notice a substantial amount of wrong ECC across the board (expected, as the model turned out not to have any training data in this category). To investigate this further we have plotted the confusion matrices for ECC in Figures \ref{fig:conf_matrix_16_08}, \ref{fig:conf_matrix_14_09}, and \ref{fig:conf_matrix_10_09} for \texttt{M1}, \texttt{M2}, and \texttt{M3} respectively. We observed that the two most popular ECCs for \emph{blogs/tech} are \emph{wiki}, and \emph{blogs} across all hyper-parameter configurations. While \emph{blogs} is a semantic supercategory of \emph{blogs/tech}, such ECC generation is most probably coincidental. Additionally, we noticed that texts in \emph{blogs/tech} have substantially more sampling loops for \texttt{M3}, indicating that the unknown control codes might make the model more conservative in its predictions, i.e., finding ``safe'' predictions and repeating them.

Moving to the categories with training data, the most problematic minor category in terms of wrong ECC is \emph{blogs/economy}, accompanied by the major categories of \emph{literature} and \emph{simple}. We also note that the larger the repetition penalty, the more cases of an erroneous  ECC we observed (which is perhaps not surprising). Looking at the ECC confusion matrices, the most frequently used ECC for \emph{blogs/economy} is \emph{news/economy}, which might be an indication that these two categories were quite similar in our training data (and/or that our automatic classifier might not have distinguished these two categories well). On the other hand, the major category \emph{simple} was most often confused with \emph{info/medical} across the board. At the same time, \emph{literature} was most frequently confused with \emph{blogs} or \emph{news/opinion} for \texttt{M1} and \texttt{M2}, and with \emph{review} for \texttt{M3}.

Regarding sampling loops, as expected, both \texttt{M1} and \texttt{M2} decreased their amount substantially across the board (see Figures \ref{fig:repeats_all_16_08} and \ref{fig:repeats_all_14_09} respectively). For texts sampled using \texttt{M3}, the number of sampling loops was substantially higher. Specifically, we observed many sampling loops for the minor categories \emph{forum/economy}, \emph{info/business}, \emph{news/tech}, \emph{news/food}, \emph{forum/travel}, \emph{forum/sport}.

Most of the time, we observed the generated texts to be lexically diverse (low BLEU-4 scores), having almost no overlap with training data. A few cases when the texts were more lexically similar to each other included texts for \emph{info/travel} with \texttt{M1} (see Figure \ref{fig:bleu_all_16_08}), for \emph{literature} and \emph{wiki} with \texttt{M2} (see Figure \ref{fig:bleu_all_14_09}). In all these cases the texts are \emph{occasional outliers}. For \texttt{M3}, the number of lexically similar texts is larger, especially for \emph{blogs/tech}, \emph{info/medical}, \emph{info/travel}. 

\clearpage
\clearpage


The 13-gram overlap between the generated texts and the training data is low for all hyper-parameter combinations, as reported in Figure \ref{fig:overlap_all}. Some categories include a small number of outlier texts with a substantial overlap: text for \emph{wiki} with \texttt{M1}, \emph{literature}, \emph{info/travel}, and \emph{info/medical} with \texttt{M2}, as well as \emph{forum/economy}, \emph{info/medical} and \emph{info/travel} for \texttt{M3}. We note that, in general, the overlap is somewhat higher for \texttt{M3}, especially for \emph{info/medical} and \emph{info/travel}. However, this is accompanied by relatively high BLEU-4 scores (see Figure \ref{fig:bleu_all_10_09}), which is an indication that the model generates slight variations of the same text, which happens to overlap with the training data. Note that a similar range of BLEU-4 scores is observed for \emph{wiki}, but the overlap with the training data for this category is much lower for \texttt{M3}, which might indicate that the ``default'' text the model generates for this category happens to be different enough from texts in the training data by chance.

Following these insights gained from the automatic evaluation metrics, we deem texts of the following categories to be of potentially very low quality: \emph{blogs/economy}, \emph{forum/economy}, \emph{info/business}, \emph{literature}, \emph{simple}. The texts of potentially low quality include \emph{news/tech}, \emph{news/food}, \emph{news/fashion}, \emph{forum/sport}, \emph{forum/travel}. The texts from the aforementioned categories were excluded from further evaluation. 

The texts of potentially relatively good quality, but with a possibility of high training data overlap include texts from the categories \emph{info/medical}, \emph{info/travel}, \emph{wiki}. The texts in all other categories could be of good quality and warrant further investigation.

\section{Human evaluation}\label{sec:human-eval}
We have conducted a small-scale human evaluation aimed at answering the following questions:
\begin{itemize}
	\item Does better performance on the automatic metrics from Section \ref{sec:gen-hp} indicate higher quality of texts?
	\item How do texts generated by SweCTRL-Mini compare to texts produced by OpenAI's GPT-3 \citep{brown2020language}?
\end{itemize}

\subsection{Text selection}\label{sec:human-text-sel}
In order to answer the above questions, we have selected categories meant to include texts written by a single author. When selecting these categories, we have created 3 buckets depending on the number of training documents in that category: (1) more than 1M, (2) between 100K and 1M, (3) between 0 and 100K. Then we have chosen a major and a minor category from each of these buckets (if possible), taking into account the insights from automatic evaluation. The final selection of categories included (with buckets specified in parentheses):
\begin{itemize}
	\item 4 major: news (1), wiki (2), ads (2), and review (3);
	\item 4 minor: news/sport (2), blogs/sport (3), news/travel (3), and info/travel (3).
\end{itemize}
For each of these categories we have devised prompts of 3 types, coded as follows:
\begin{itemize}
	\item \textbf{names}, talking about people/objects with previously unseen names;
	\item \textbf{future}, talking about events happening in future compared to the time of writing this article (later than 2023);
	\item \textbf{existing}, taking a beginning of the textual material from the training data of SweCTRL-Mini for the given category.
\end{itemize}
The rationale behind the first two prompt types was to let the model generate text in domains that have most probably not been included in its training data (although one can not be sure with closed-source models such as GPT-3). For these two types, we have devised 1 prompt for each. The reason for including the last prompt type was to get an indication of overfitting. We have devised 2 prompts for this type, bringing the total number of prompts per each tested category to 4.

We sampled 3 times for each pair of prompt (one of 4 types mentioned above) and set of hyper-parameters (one of \texttt{M1}, \texttt{M2}, or \texttt{M3} for SweCTRL-Mini, or the default set of hyper-parameters for GPT-3, described in Appendix \ref{app:gpt3-setup}). Hence, in total, $8 \cdot 4 \cdot 3 \cdot 4 = 384$ texts were subject to human evaluation.

\subsection{Setup}
The English translation of the guidelines is provided in Figure \ref{fig:en-guidelines}, and the original guidelines in Swedish are provided in the technical note \citep[Section 4]{kalpakchi_dmytro_2023_7868205}. To give a brief summary, we asked human judges to find specific kinds of errors, and mark only the first occurrence of each kind in the text (if they occur). The types of errors we were interested in are: (1) stylistic errors, (2) topic shifts, (3) grammatical errors, (4) vocabulary errors, and (5) factual errors. Additionally, the annotators had access to a free-text field, where they could specify any other kinds of errors.

The evaluation was performed by 2 annotators (ourselves) using the instance of Textinator \citep{kalpakchi2022textinator} as the annotation tool. The total of 384 texts were divided equally on a per-prompt basis. The texts were split in groups, such that the texts in one group were generated for one prompt by all given models. The order of texts within each group was randomized and we did not know which text belonged to which model. The order of groups was also randomized. In order to recover the model information, we generated a separate key file, which we did not access until we were done annotating. The annotation was done using an iterative process (annotating -- discussing issues -- re-annotating) and all annotations will be made publicly available in the associated GitHub repository, in addition to the supplied analysis.

\begin{figure*}
	\begin{tcolorbox}
		Thank you for your help! You will see a bunch of texts (one at a time). The author of each text has been instructed to perform a specific task, for instance:
		\vspace{0.7em}
		
		\textit{Write a blog post starting with a phrase "Stockholm is best at"}
		\vspace{0.7em}
		
		All tasks were of the same kind: the writer was asked to write a text in a specific genre (blog post in the example above), which starts with a specific starting phrase ("Stockholm is best at" in the example above).
		\vspace{0.7em}
		
		\textbf{YOUR TASK} is to review texts and find the following kinds of errors (if any):
		\begin{enumerate}
			\item \textbf{stylistic errors}, when the student wrote in a different genre, for instance, a Wikipedia article, instead of a blog post.
			\item \textbf{Topic shifts}, when the text is largely unrelated to the starting phrase.
			\item \textbf{Grammatical errors}, when there are mistakes in spelling (``\textit{\underline{Sockholm} is the capital of Sweden}''), word order (``\textit{\underline{Have done you} your homework?}''), or wrong inflection (``\textit{I \underline{will go} to the cinema for two days ago}'').
			\item \textbf{Vocabulary errors}, when the writer used a word/expression in a wrong way (for instance, ``\textit{The Earth jumps around The Sun}'', or ``\textit{Clapping with my arms is so fun!}'').
			\item \textbf{Factual errors}, when the writer stated facts, which constitute general knowledge, in a wrong way (for instance, ``\textit{1 kilogram equals to 30 grams}'', or ``\textit{The Sun rotates around The Earth}'', or ``\textit{Dogs typically fly very fast}'').
		\end{enumerate}
		
		Should you find \textbf{an error of the kind 1 or 2}, please tick the appropriate checkbox on the ``Markers'' panel to the right.
		\vspace{0.7em}
		
		Should you find \textbf{an error of the kind 3, 4, or 5}, please use the mouse to mark the place in the text where you have found the error, and then click on the corresponding marker on the panel ``Markers'' to the right.
		\vspace{0.7em}
		
		The texts can finish abruptly, this should \textbf{NOT} be viewed as an error.
		\vspace{0.7em}
		
		Should you find any other kind of errors, please briefly note your findings in the field named ``Other errors''.
		\vspace{0.7em}
		
		You do \textbf{NOT} need to find all errors of all kinds! If you find more than 1 error of any kind, for instance, grammatical errors, it is enough to mark only the first occurrence. However, if you have found 1 vocabulary error and 1 factual error, you should mark both, because they are different kinds of errors.
		\vspace{0.7em}
		
		If you couldn't find any errors, please tick the checkbox under ``Fully correct'' on the ``Markers'' panel to the right.
		\vspace{0.7em}
		
		When you are done with the text, please click on the ``Get new text'' button in the bottom right corner of the screen.
	\end{tcolorbox}
	\caption{The approximate English translation of the instructions for human evaluation (see technical note \cite[Section 4]{kalpakchi_dmytro_2023_7868205} for the original guidelines in Swedish)}
	\label{fig:en-guidelines}
\end{figure*}

\subsection{Results}
\begin{figure*}[!htb]
	\centering
	\includegraphics[width=\textwidth]{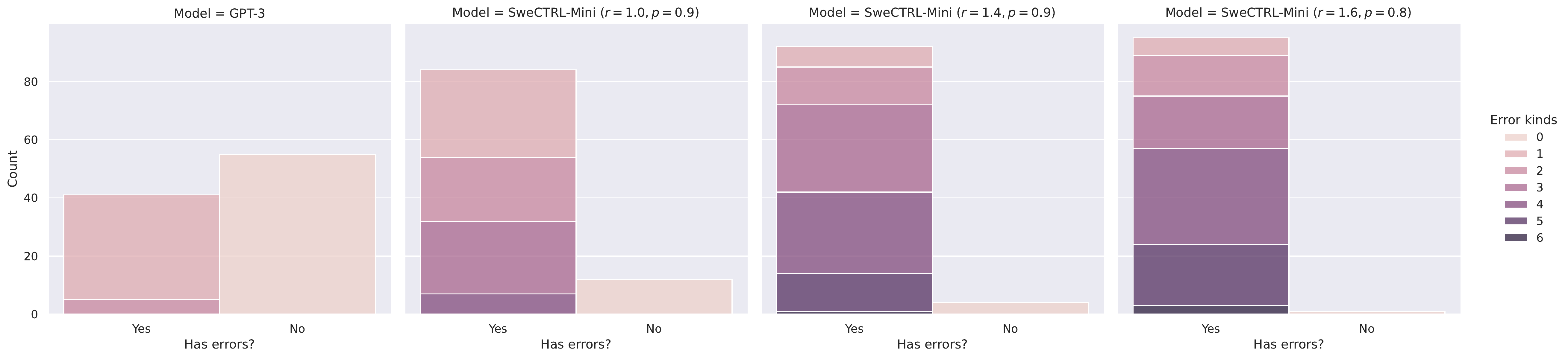}
	\vspace{-1.2em}
	\caption{The distribution of texts by the number of error kinds for models evaluated by human judges}
	\label{fig:human_has_errors}
\end{figure*}

Recall that 384 texts were subject to human evaluation for all 4 models, making it 96 texts per model. As reported in Figure \ref{fig:human_has_errors}, 55 out of 96 texts (57.29\%) produced by GPT-3 were error-free, which is substantially more texts with no errors compared to all versions of SweCTRL-Mini. Among the versions of SweCTRL-Mini, the best model was using hyper-parameters from \texttt{M3} (contrary to automatic evaluation) and managed to generate only 12 error-free texts (12.5\%), which is still substantially larger than the other versions of SweCTRL-Mini.

\begin{figure*}[!htb]
	\centering
	\includegraphics[width=\textwidth]{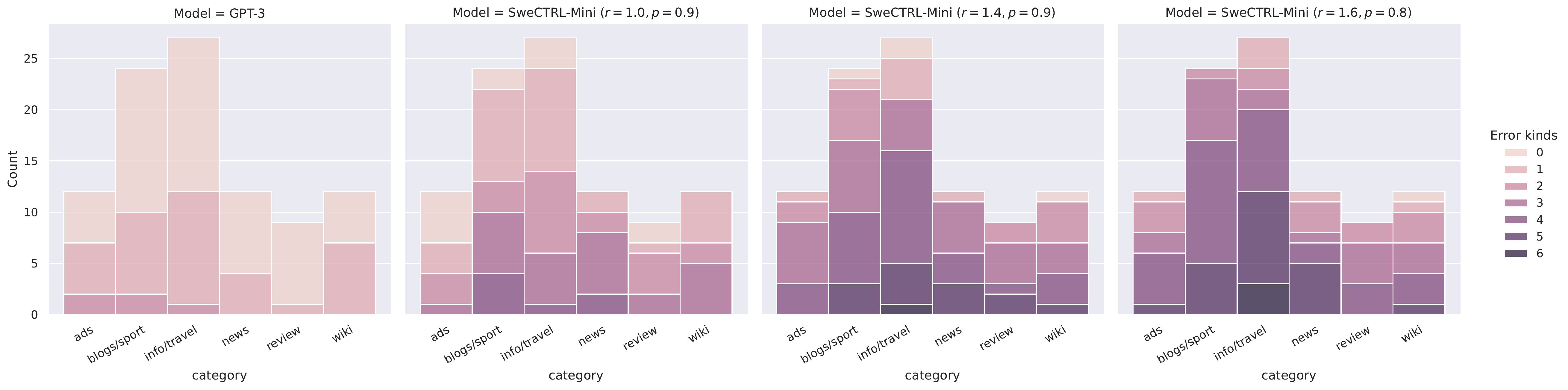}
	\vspace{-1.2em}
	\caption{The distribution of the number of error kinds by category for models evaluated by human judges}
	\label{fig:human_num_errors_cat}
\end{figure*}

\begin{figure*}[!htb]
	\centering
	\includegraphics[width=\textwidth]{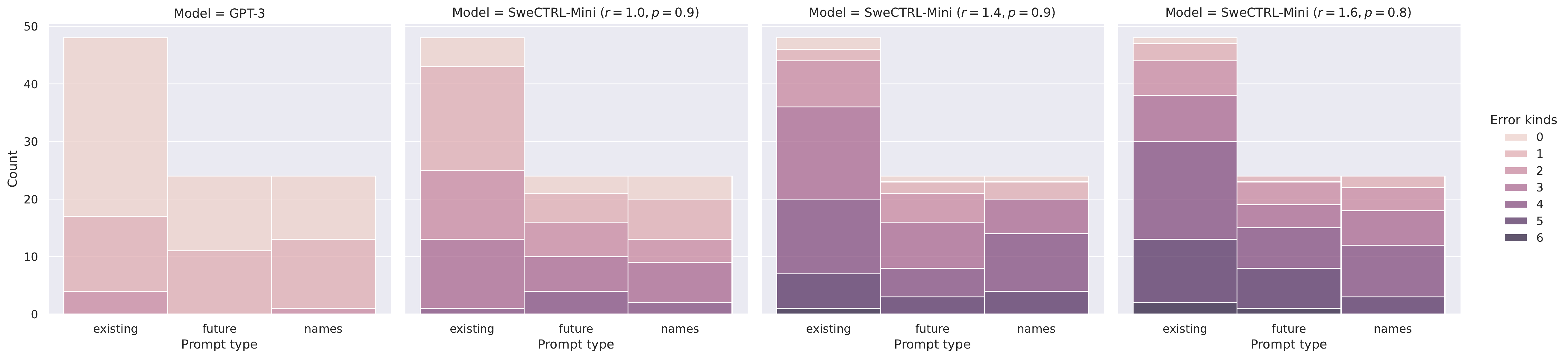}
	\vspace{-1.2em}
	\caption{The distribution of the number of error kinds by the prompt type for models evaluated by human judges}
	\label{fig:human_num_errors_prompt}
\end{figure*}

Looking closely at the distribution of the number of error kinds per category in Figure \ref{fig:human_num_errors_cat}, we observed that the only category where texts produced by GPT-3 and SweCTRL-Mini had a comparable number of errors is the category \emph{ads} for \texttt{M3}. We also note that the larger the repetition penalty $r$, the more error kinds have been found by human judges. Similar trends could be observed per prompt type in Figure \ref{fig:human_num_errors_prompt}.

\begin{figure*}[!htb]
	\centering
	\includegraphics[width=\textwidth]{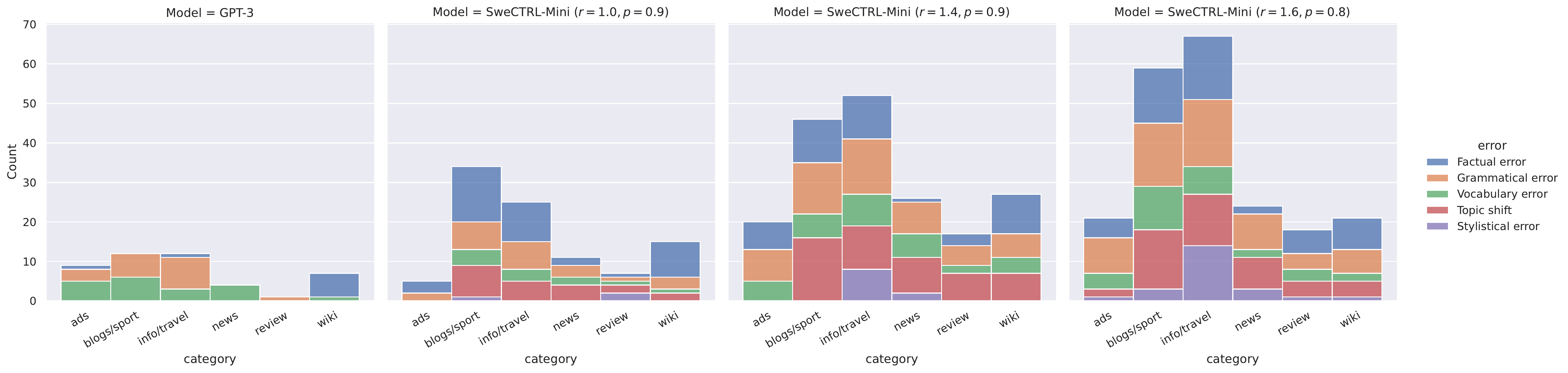}
	\vspace{-1.2em}
	\caption{The distribution of main error kinds by category for models evaluated by human judges}
	\label{fig:human_cat_main_errors}
\end{figure*}

\begin{figure*}[!htb]
	\centering
	\includegraphics[width=\textwidth]{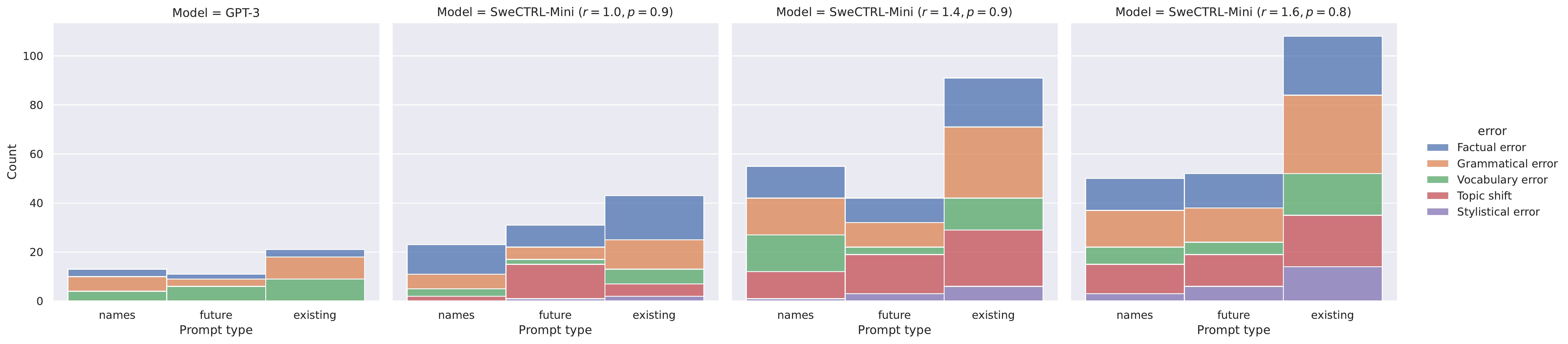}
	\vspace{-1.2em}
	\caption{The distribution of main error kinds by the prompt type for models evaluated by human judges}
	\label{fig:human_prompt_main_errors}
\end{figure*}

Looking closer at the specific kinds of errors, we have divided the errors into two categories: {\it main} (explicitly specified by the guidelines) and {\it other} (the ones distilled from the text field comments). The per-category distribution of the main error kinds is presented in Figure \ref{fig:human_cat_main_errors} (note that one text can have more than 1 error kind). We observed a number of trends:
\begin{itemize}
	\item texts produced by GPT-3 did not have any stylistic errors or topic shifts;
	\item the number of stylistic errors and topic shifts increased with the repetition penalty~$r$ for SweCTRL-Mini;
	\item all versions of SweCTRL-Mini had substantially more factual errors than GPT-3, except for the category \emph{wiki}, where they were roughly equal;
	\item the number of grammatical errors was comparable between GPT-3 and SweCTRL-Mini using \texttt{M3}, while substantially larger for \texttt{M1}, and \texttt{M2};
	\item the number of vocabulary errors was the lowest for SweCTRL-Mini using \texttt{M3} followed by GPT-3, \texttt{M2}, and \texttt{M1}.
\end{itemize}
Similar trends are observed in the per-prompt-type distribution of main error kinds, shown in Figure \ref{fig:human_prompt_main_errors}. Here we note that most topic shifts happened for the prompts related to the future for \texttt{M3}, and were much rarer for the other prompt types. The other two versions of SweCTRL-Mini (with $r > 1$) had comparable numbers of topic shifts across the board.

\begin{figure*}[!htb]
	\centering
	\includegraphics[width=\textwidth]{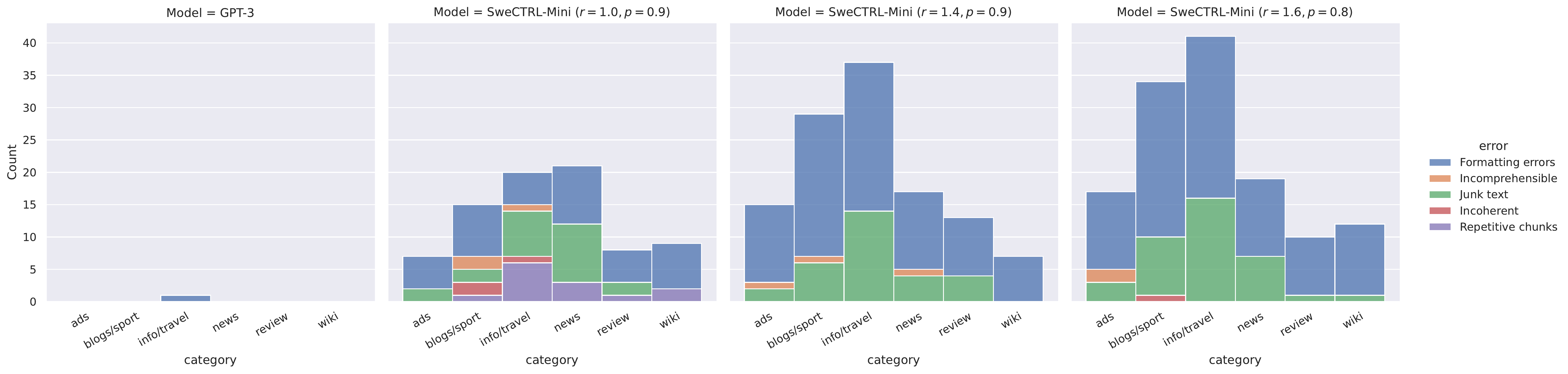}
	\vspace{-1.2em}
	\caption{The distribution of other error kinds by category for models evaluated by human judges}
	\label{fig:human_cat_other_errors}
\end{figure*}

\begin{figure*}[!htb]
	\centering
	\includegraphics[width=\textwidth]{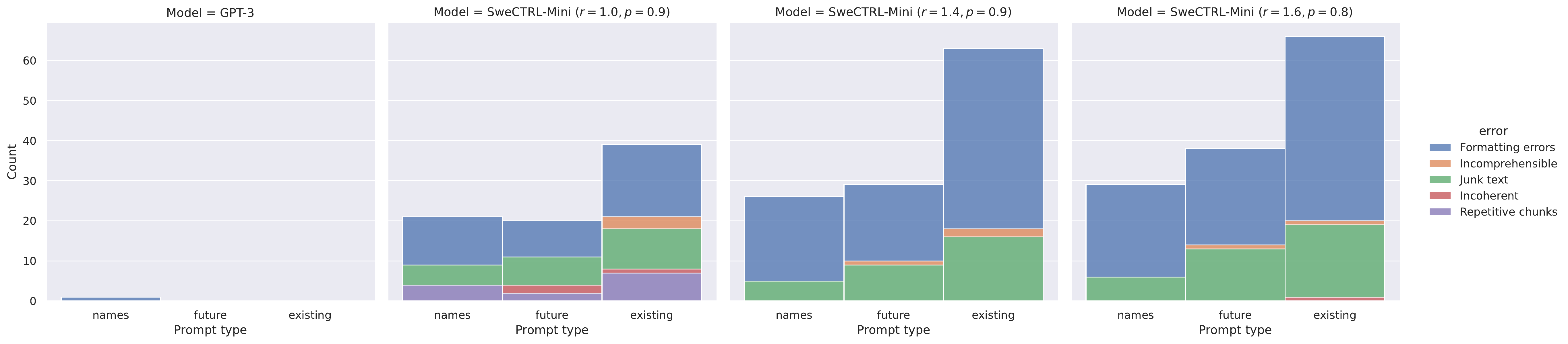}
	\vspace{-1.2em}
	\caption{The distribution of other error kinds by the prompt type for models evaluated by human judges}
	\label{fig:human_prompt_other_errors}
\end{figure*}

The distribution of the other error kinds is presented in Figure \ref{fig:human_cat_other_errors} per category and in Figure \ref{fig:human_prompt_other_errors} per prompt type (note that one text can have more than 1 error kind). We observed a number of trends:
\begin{itemize}
	\item Almost no texts produced by GPT-3 had any additional specified problems, except one formatting error. By contrast, formatting errors were the largest identified kind of problem for SweCTRL-Mini;
	\item All versions of SweCTRL-Mini produced a substantial number of junk text (artifacts that we failed to clean from the training data, e.g., ``Likes 89'', ``Log in Logout'', or ``Photo: Erik Johnson'') across the categories and prompt types;
	\item As expected, the number of repetitive chunks decreases as the repetition penalty $r$ increases, similar to the findings of automatic evaluation;
	\item A number of texts produced by SweCTRL-Mini were marked as incomprehensible or incoherent. The number is very small and did not seem to be particular to any specific version of the model, category, or prompt type.
\end{itemize}

\section{Evaluation on SuperLim}
SuperLim \citep{adesam2020swedishglue} is a standardized benchmark for evaluating NLI systems in Swedish, comprised of 15 tasks. We attempted to fine-tune SweCTRL-Mini on 11 tasks from SuperLim 2.0.4 that are either on the text or the sentence level (we will later refer to such tasks as \emph{TS-tasks}). These tasks are briefly summarized in Table \ref{tab:desc_superlim2}. For the purpose of this work, we have used only training and test sets. The overlaps $O_*^7$ and $O_*^{13}$ between these sets and the training data of SweCTRL-Mini are reported in Table \ref{tab:superlim2_overlap}.

\subsection{Setup}
\begin{table*}[!htb]
	\centering
	\begin{tabular}{p{2.6cm}p{5cm}cccc}
		\toprule
		\textbf{Dataset} & \textbf{Task description} & \textbf{$|$Tr$|$} & \textbf{$|$D$|$} & \textbf{$|$T$|$} & \textbf{Ver}\\
		\midrule
		ABSAbank-Imm (abbr. A-Imm) & Given a text, label the expressed sentiment towards immigration in Sweden & 3898 & 487 & 487 & 1.1 \\
		\midrule
		SweDN  & Given a text from a news article, provide its summary & 29.8K & 4529 & 3745 & 1.0\\
		\midrule
		SweWinograd (abbr. SweWgd) & Check if the pronoun and candidate antecedent co-refer  & 339 & 55 & 43 & 2.0 \\
		\midrule
		SweFraCas (abbr. SweFC) & Given a yes/no question and a number of premises, choose the suitable answer & \redcross & \redcross & 305 & 1.0 \\
		\midrule
		\midrule
		DaLAJ-GED (short DaLAJ) & Check if a given sentence is grammatically correct & 35.5K & 4702 & 4371 & 2.0\\
		\midrule
		SweNLI & Check if a premise entails a hypothesis. & 393K & 9815 & 305 & 1.0 \\
		\midrule
		SweFAQ & Given the question, find the suitable answer & 9603 & 1452 & 2007 & 2.0 \\
		\midrule
		SweParaphrase (short SwePara) & Given two sentences determine how similar they are & 5715 & 1499 & 1378 & 2.0 \\
		\midrule
		SweWiC & Predict if the uses of the word in two different contexts constitute the same sense. & 4486 & 500 & 1000 & 2.0\\
		\midrule
		SweWinogender (abbr. SweWgr) & Check if a discourse fragment with a pronoun entails a sentence with the pronoun replaced by a candidate antecedent & \redcross & \redcross & 624 & 2.0\\
		\midrule
		SweDiagnostics (short SweDia) & Check if a premise entails a hypothesis. & \redcross & \redcross & 1104 & 1.1 \\
		\bottomrule
	\end{tabular}
	\caption{\label{tab:desc_superlim2} Dataset descriptions for TS-tasks in SuperLim 2.0.4. ``Tr'' stands for ``training set'', ``D'' stands for ``development set'', ``T'' stands for ``test set'', K stands for ``thousands''.}
\end{table*}

\begin{table*}[!htb]
	\centering
	\begin{tabular}{p{1.2cm}p{0.2cm}cccccccc}
		\toprule
		\multirow{2}{*}{\textbf{Dataset}} & & \multicolumn{4}{c}{7-grams} & \multicolumn{4}{c}{13-grams} \\
		& & \textbf{$N_{<7}$} & \textbf{$O_1^7$} & \textbf{$O_{10}^7$} & \textbf{$O_{100}^7$} &  \textbf{$N_{<13}$} & \textbf{$O_1^{13}$} & \textbf{$O_{10}^{13}$} & \textbf{$O_{100}^{13}$} \\
		\midrule
		A-Imm & Tr & 6.9\% & 28.25\% & 0.49\% & 0.06\% & 19.2\% & 26.36\% & 0.02\% & 0\% \\
		& T & 4.9\% & 28.47\% & 0.49\% & 0.08\% & 22.0\% & 27.12\% & 0\% & 0\% \\
		\midrule
		SweDN & Tr & 0\% & 17.79\% & 0.65\% & 0.11\% & 0\% & 14.33\% & 0.04\% & XS \\
		& T & 0\% & 27.69\% & 0.47\% & 0.06\% & 0\% & 24.76\% & 0.02\% & XS \\
		\midrule
		SweWgd & Tr & 0\% & 0.46\% & 0\% & 0\% & 24.6\% & 0\% & 0\% & 0\% \\
		& T & 0\% & 0.13\% & 0\% & 0\% & 1.4\% & 0\% & 0\% & 0\% \\
		\midrule
		SweFC & T & 53.9\% & 0.17\% & 0.17\% & 0 & 95.6\% & 0\% & 0\% & 0\% \\
		\midrule
		\midrule
		DaLAJ & Tr & 10.8\% & 2.79\% & 0.44\% & 0.08\% & 40.6\% & 0.20\% & XS & 0 \\
		& T & 11.6\% & 2.73\% & 0.49\% & 0.11\% & 39.4\% & 0.02\% & 0 & 0 \\
		\midrule
		SweNLI & Tr & 21.1\% & 0.45\% & 0.09\% & 0.02\% & 60.2\% & \multicolumn{3}{c}{all three XS}\\
		& T & 35.4\% & 0.10\% & 0.10\% & 0\% & 84.1\% & 0\% & 0\% & 0\% \\
		\midrule
		SweFAQ & Tr & 8.5\% & 15.10\% & 0.90\% & 0.11\% & 34.5\% & 12.36\% & 0.20\% & 0\%\\
		& T & 7.7\% & 10.05\% & 0.46\% & XS & 40.6\% & 5.66\% & 0\% & 0\%\\
		\midrule
		SwePara & Tr & 35.1\% & 4.27\% & 0.04\% & XS & 80.6\% & 0\% & 0\% & 0\% \\
		& T & 38.0\% & 2.02\% & 0.65\% & 0.15\% & 79.8\% & 0\% & 0\% & 0\% \\
		\midrule
		SweWiC & Tr & 30.2\% & 3.58\% & 0.33\% & 0.15\% & 79.5\% & 3.23\% & 0\% & 0\% \\
		& T & 5.6\% & 13.92\% & 0.28\% & 0.03\% & 27.4\% & 12.2\% & 0\% & 0\% \\
		\midrule
		SweWgr & T & 11.5\% & 0.10\% & 0\% & 0\% & 74.0\% & 0\% & 0\% & 0\% \\
		\midrule
		SweDia & T & 15.1\% & 0.48\% & 0.06\% & 0\% & 46.2\% & 0\% & 0\% & 0\% \\
		\bottomrule
	\end{tabular}
	\caption{\label{tab:superlim2_overlap} Overlap in 7-grams and 13-grams between the datasets for TS-tasks and the training data of SweCTRL-Mini. ``Tr'' stands for ``training set'', ``T'' stands for ``test set''. $N_{<k}$ denotes the percentage of texts in the dataset $D$ that was shorter than $k$ tokens (after splitting by spaces). XS denotes non-zero values $<$0.01\%.}
\end{table*}

For the 8 TS-tasks that provide training data, we have fine-tuned \emph{all weights} of SweCTRL-Mini on the entire provided training data. For every task, we have viewed each datapoint as having 2 components: the prompt \texttt{P} (consisting of a text and a criterion of interest), and a label \texttt{L} providing the solution to the problem in \texttt{P}. Additionally, we introduced special OCC and ECC for each task. All of these components needed to fit into the context window of 256 tokens, which is a limitation of our model. To achieve that, we introduced a hard limit $H_{P}$ on the number of tokens that could be spent on \texttt{P}, which was calculated as follows:
\begin{equation*}
	H_{P} = \begin{cases}
		245, \text{if } |\texttt{P}| + |\texttt{L}| + 2 \le 256 - \delta,\\
		256 - \delta - |\texttt{L}| - 2, \text{otherwise}
	\end{cases}
\end{equation*}
For cases when the prompt turned out to be too long, i.e., $|\texttt{P}| > H_{P}$, we have split it by taking $\big\lfloor\frac{H_{P}}{2}\big\rfloor$ tokens from the beginning and the end of \texttt{P}. In order to indicate that there was a piece of text in the middle, we introduced an additional separator string \texttt{[...]} in between them, which we required to be tokenized as at most $\delta = 5$ tokens\footnote{In practice it turned out to be only 1 token, but $\delta$ remained to be 5 for the rest of the experiments}.

The vast majority of prompts have been formulated as question-answering tasks requiring the answer ``Yes'', ``No'', or ``Maybe'', except for tasks requiring to provide a numeric score (ABSAbank-Imm, SweParaphrase),  or a summary (SweDN). Each datapoint of any SuperLim task was represented as the following string:
\begin{center}
\begin{BVerbatim}
[OCC] [Prompt] [Label] [ECC]
\end{BVerbatim}
\end{center}
To provide an example, an instance of SweDN task, would be converted into:
\begin{center}
	\begin{tabular}{llll}
		\texttt{[OCC]} & \texttt{[Prompt]} & \texttt{[Label]} & \texttt{[ECC]} \\
		\texttt{:swedn:} & \texttt{[Text] Sammanfattning:} & \texttt{[Summary]} & \texttt{:swedn:\$}
	\end{tabular}
\end{center}

\noindent The exact prompt formulations used for each SuperLim task are provided in Table \ref{tab:superlim2_prompts}.

\begin{table*}[!htb]
	\centering
	\begin{tabular}{lp{9.5cm}}
		\midrule
		\textbf{Dataset} & \texttt{[Prompt] [Label]}\\
		\midrule
		ABSAbank-Imm & \texttt{T} Känsloläge: [Rating]\\
		SweDN & \texttt{T} Sammanfattning: [Summary]\\
		SweWinograd & \texttt{T} Fråga: Syftar '\texttt{W1}' till '\texttt{W2}'? Svar: [Ja/Nej]\\ 
		SweFraCas & Premiss: \texttt{P} Fråga: \texttt{Q} Svar: [Ja/Nej/Vet ej/Jo]\\
		\midrule
		DaLAJ-GED & \texttt{T} Fråga: Är meningen grammatiskt korrekt? [Ja/Nej] \\
		SweNLI & Situation: \texttt{P} Påstående: \texttt{H} Fråga: Stämmer? Svar: [Ja/Nej/Kanske] \\
		SweFAQ & Fråga: \texttt{Q} Svar: \texttt{A} Passar? [Ja/Nej] \\
		SweParaphrase & Mening 1: \texttt{S1} Mening 2: \texttt{S2} Likhet mellan meningar: [Rating] \\
		SweWiC & Text 1: \texttt{T1} Text 2: \texttt{T2} Fråga: Betyder ordet '\texttt{W}' samma sak i båda fall? Svar: [Ja/Nej]\\
		SweWinogender & Situation: \texttt{P} Påstående: \texttt{H} Fråga: Stämmer? [Ja/Nej/Kanske]\\
		SweDiagnostics & Situation: \texttt{P} Påstående: \texttt{H} Fråga: Stämmer? [Ja/Nej/Kanske]\\
		\bottomrule
	\end{tabular}
	\caption{Prompts used for fine-tuning on TS-tasks in SuperLim 2.0.4. The text in [square brackets] indicates the label to be predicted.}\label{tab:superlim2_prompts}
\end{table*}

% TODO: remove info about random seeds
For all experiments, we fine-tuned SweCTRL-Mini for 1, 3, and 5 epochs using the batch size of 4 on one NVIDIA 3090 GPU with 24GB of VRAM. We used the AdamW optimizer \citep{loshchilov2018decoupled} with the initial learning rate \num{5e-5}, $\beta_1 = 0.9, \beta_2 = 0.999, \epsilon = \num{1e-8}$. The gradients were clipped to the norm of 1. Additionally, at the beginning of fine-tuning for each fold, we have fixed the random seed to a prime number (specifically, 87178291199), in order to initialize embeddings for the task-dependent OCC and ECC in the same way.

Importantly, when evaluating, we have employed \textbf{greedy} generation by taking the token with the maximal probability. In order to avoid cases when the model chooses the generate the ECC directly, we have prohibited the first generated token to be the ECC token corresponding to the SuperLim task at hand.

For each SuperLim task except SweDN, we also provide a baseline majority-vote performance on the test set using the most frequent class of \emph{the very same test set} (not the training set). In a real-world scenario, implementing such a baseline would be impossible, because the test set is typically not available. However, in the case of benchmarks, especially as recent as SuperLim 2, such a baseline provides a necessary reference point to reflect on the performance of SweCTRL-Mini. For SweDN, composed of news articles, we take the first sentence as a summary using the sentence splitter from Stanza 1.5.0 \citep{qi2020stanza}, and convert the obtained dependency trees to strings using UDon2 \citep{kalpakchi2020udon2}.

\subsection{Evaluation metrics}
For all tasks, we have used the evaluation metrics suggested by SuperLim developers. Here we explain our interpretation of these metrics.

\textbf{Krippendorff's alpha} $\alpha$ introduced by \citet{krippendorff1970bivariate} to measure the inter-annotator agreement is a number one choice for many tasks in SuperLim. We use this metric by assuming that ground truth labels have been provided by annotator~1, whereas predictions by a fine-tuned SweCTRL-Mini have been provided by annotator~2. If SweCTRL-Mini did not manage to provide a meaningful prediction for the task (e.g., failing to predict a number, when a score is required), we deem such a case as a missing annotation and exclude it from both annotators 1 and 2. The statistics on missing annotations are reported separately. For all tasks, alpha was calculated using NLTK version 3.8.1 \citep{loper2002nltk}.

\textbf{Kripendorff's pseudoalpha} $\alpha^+$ calculated only for one task, SweFAQ, using the following formula:
$$\alpha^+ = \frac{(A - \frac{109}{2049})}{\frac{1940}{2049}},$$
where $A$ is the proportion of questions correctly connected to their answer.

\textbf{Spearman's rank correlation coefficient} $\rho$ introduced by \citet{spearman1904} was used in the same way as Krippendorff's alpha.

\subsection{Results}
The performance of the fine-tuned SweCTRL-Mini models is summarized in Table \ref{tab:superlim2_full_train_res}. As can be seen, SweCTRL-Mini manages to beat the baseline on 5 out of 8 tasks, sometimes already after 1 epoch of fine-tuning (for ABSAbank-Imm, DaLAJ-GED, and SweParaphrase), and sometimes requiring more epochs (3 epochs for SweNLI, SweWiC). SweCTRL-Mini did not manage to beat the baseline for 2 tasks: SweWinograd, and SweFAQ. While for SweWinograd the performance was steadily increasing with more training, it is likely that more fine-tuning will lead to better results. By contrast, the best results for SweFAQ were achieved after 1 epoch with performance decreasing down the line (thus indicating overfitting). 

\begin{table*}[!htb]
	\centering
	\begin{tabular}{lp{2cm}cccc}
		\midrule
		\textbf{Dataset} & \textbf{Metric} & \textbf{Base} & \textbf{E1} & \textbf{E3} & \textbf{E5}\\
		\midrule
		ABSAbank-Imm & $\alpha$ & -0.2197 & 0.2218 & 0.2747 & 0.2935 \\
		& $\rho$ & NaN & 0.4669 & 0.5308 & 0.5046 \\
		& $N_{m}$ & 0\% & 10.06\% & 26.49\% & 29.35\%\\
		\midrule
		SweDN & ROUGE-L & \multicolumn{4}{c}{See Figure \ref{fig:swedn_rouge}}\\
		\midrule
		SweWinograd & $\alpha$ & -0.1772 & -0.6431 & -0.3728 & -0.2546 \\
		& $N_{m}$ & 0\% & 0\% & 0\% & 0\%\\
		\midrule
		\midrule
		DaLAJ-GED & $\alpha$ & -0.3264 & 0.6514 & 0.2415 & -0.1418 \\
		& Accuracy & 0.5076 & 0.8213 & 0.5150 & 0.2391 \\
		& $N_{m}$ & 0\% & 0\% & 0\% & 0\%\\
		\midrule
		SweNLI & $\alpha$ & -0.19707 & -0.2473 & -0.1287 & -0.0551 \\
		& Accuracy & 0.5934 & 0.1639 & 0.2065 & 0.3213 \\
		& $N_{m}$ & 0\% & 0\% & 0\% & 0\%\\
		\midrule
		SweFAQ & $\alpha^+$ & 0.9426 & 0.9411 & 0.9137 & 0.5785 \\
		& Accuracy & 0.9457 & 0.9442 & 0.9183 & 0.6009 \\
		\midrule
		SweParaphrase & $\alpha$ & -0.3137 & -0.0206 & 0.0135 & 0.0446 \\
		& $N_{m}$ & 0\% & 0.58\% & 1.35\% & 10.81\%\\
		\midrule
		SweWiC & Accuracy & 0.5 & 0.403 & 0.534 & 0.592 \\
		\bottomrule
	\end{tabular}
	\caption{\label{tab:superlim2_full_train_res} Results of fine-tuning SweCTRL-Mini for TS-tasks of SuperLim 2.0.4 on the full training sets (for the datasets where it was available). $\alpha$ denotes Krippendorff's alpha, $\alpha^+$ -- Krippendorff's pseudoalpha, $N_{m}$ -- \% of missing predictions.}
\end{table*}

\begin{figure}[!htb]
	\centering
	\includegraphics[width=0.5\linewidth]{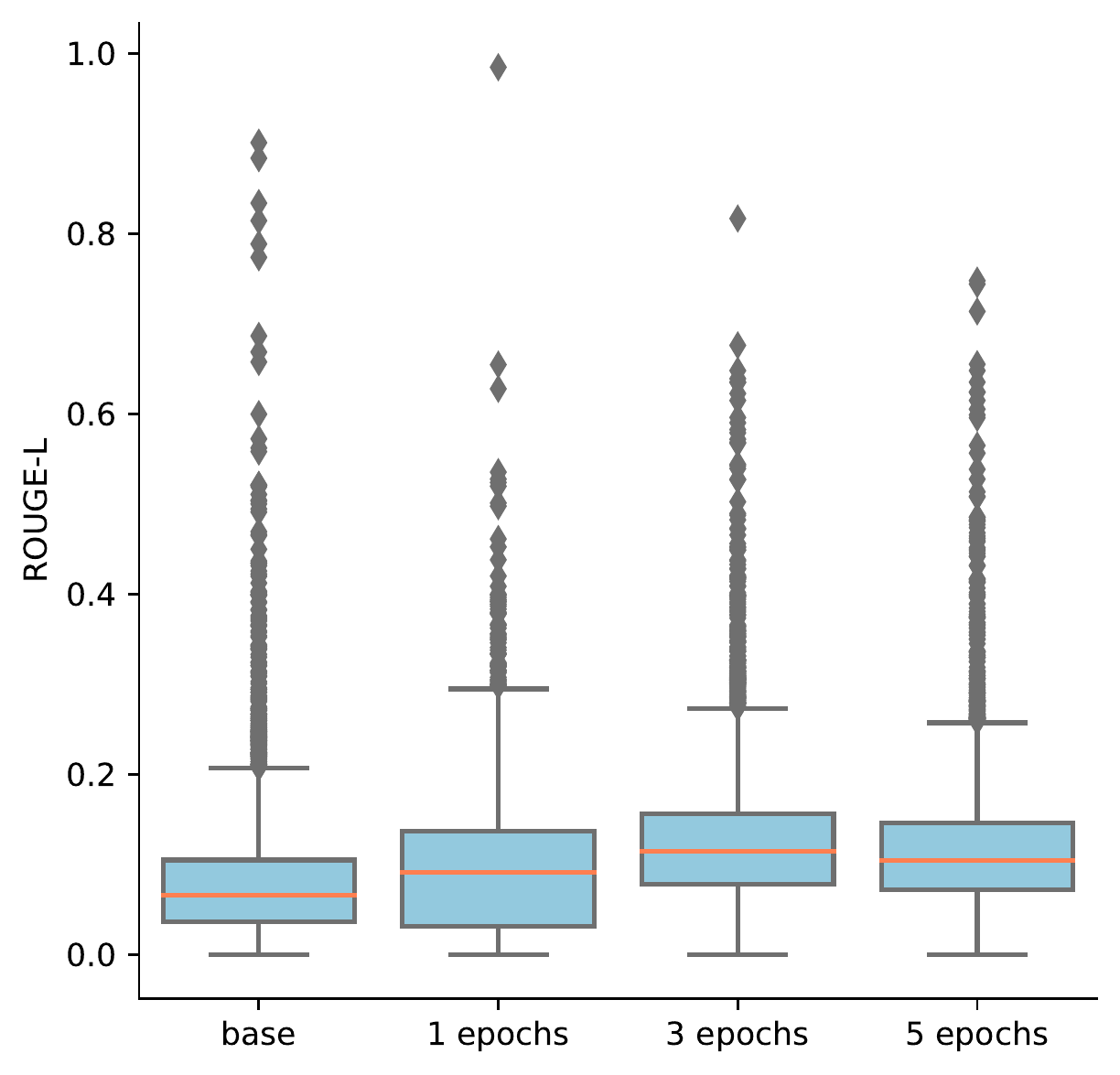}
	\caption{ROUGE-L scores for the fine-tuned SweCTRL-Mini}
	\label{fig:swedn_rouge}
\end{figure}

The results for SweDN are presented in Figure \ref{fig:swedn_rouge}. We observed that ROUGE-L scores are comparable between the epochs and that the maximum ROUGE-L score of around 1 was achieved already after 1 epoch. The median performance of all epochs outperforms the baseline, but the performance difference does not seem to be significant. That said, SweCTRL-Mini is inherently at a disadvantage with the task of text summarization due to its small context window and thus an inability to catch all the text (potentially missing parts crucial for the summary). Hence, low ROUGE-L scores (which might indicate bad quality of summary) are expected.

\section{Conclusion}\label{sec:conclusion}
In this article, we have presented SweCTRL-Mini, a Transformer-based model for Swedish allowing both generation and fine-tuning on a single consumer-grade GPU. We have observed that generated texts have generally more problems than those generated by GPT-3, although in some cases (for \texttt{M3} and ads category), the quality of texts is on-par. Among the different versions of SweCTRL-Mini we tested, there seems to be a trend that higher values of the repetition penalty $r$ lead to lower quality texts according to human evaluation (more topic shifts, stylistic, and grammatical errors). At the same time, we have observed $r=1.0$ (\texttt{M3}) to produce repeated chunks of text more frequently than $r = 1.4$ (\texttt{M2}) or $r = 1.6$ (\texttt{M1}). The observations for other hyper-parameters (nucleus threshold $p$, and temperature $T$) were less conclusive.

Another crucial observation is that SweCTRL-Mini is substantially worse at generating well-formatted texts than GPT-3. This is a clear indication that the filtering and cleaning procedures that we have employed were not enough and require revision.

When generating text greedily, we have observed that SweCTRL-Mini beats the majority vote baseline for 3 SuperLim tasks already after 1 epoch of fine-tuning (and for 5 tasks after 3 epochs)! The performance on 2 SuperLim tasks was worse than baseline which could be due to the exact formulation of the prompt. We did not experiment with different prompt formulations or any further analysis on SuperLim tasks and leave this to experts in each specific area.

\backmatter

\bmhead{Supplementary information}
This article is accompanied by a technical note that can be accessed at \url{https://doi.org/10.5281/zenodo.7868205}.

\bmhead{Acknowledgments}
The model training was enabled by the supercomputing resource BerzeLiUs provided by National Supercomputer Centre at Link\"oping University and the Knut and Alice Wallenberg Foundation. The work carried out in this project is part of the Digital Futures project ``SWE-QUEST''.

\section*{Statements and Declarations}

Some journals require declarations to be submitted in a standardised format. Please check the Instructions for Authors of the journal to which you are submitting to see if you need to complete this section. If yes, your manuscript must contain the following sections under the heading `Declarations':

\begin{itemize}
\item Funding: Digital Futures, project ``SWE-QUEST'', and Vinnova project 2019-02997.
\item Competing interests: the authors declare none.
\item Availability of data and materials: the training data is a part of publicly available corpora.
\item Code availability: the code is available at \url{https://github.com/dkalpakchi/SweCTRL-Mini}.
\item Authors' contributions: Dmytro Kalpakchi was responsible for training SweCTRL-Mini (including software implementation), designing evaluation methodology, annotating half of the texts for human evaluation, analyzing and reporting results and writing the article. Johan Boye was annotating half of the texts for human evaluation, editing the article, and providing guidance for issues arising during training and evaluation of SweCTRL-Mini. 
\end{itemize}

\begin{appendices}

\section{Setup for human evaluation of GPT-3}
\label{app:gpt3-setup}
When generating texts with GPT-3, we have used the following hyper-parameters:
\begin{itemize}
    \item temperature of 0.7;
    \item ``top p'' (for nucleus sampling) of 1;
	\item frequency and presence penalties of 0;
	\item ``best of'' being equal to 1;
	\item no custom stop sequences;
	\item maximum length of 256 tokens (to be on par with SweCTRL-Mini).
\end{itemize}

\begin{table*}[!tb]
	\centering
	\begin{tabular}{lp{5cm}p{5cm}}
		\midrule
		\textbf{Category} & \textbf{Prompt (Swedish)} & \textbf{Prompt (English)} \\
		\midrule
	    news & Skriv en nyhetsartikel. & Write a news article.\\
        wiki & Skriv en artikel i Wikipedia. & Write an article in Wikipedia. \\
        ads & Skriv en annons. & Write an advertisement.\\
        review & Skriv en recension. & Write a review.\\
        info/travel & Skriv en informerande text om resor. & Write an informative text about travelling.\\
        blogs/sport & Skriv ett blogginlägg om idrott. & Write a blog post about sport.\\
		\bottomrule
	\end{tabular}
	\caption{Initial prompts for GPT-3} \label{tab:gpt3_prompts}
\end{table*}
For each generation with GPT-3 we have used the initial prompts (in Swedish) from Table \ref{tab:gpt3_prompts}. Each such initial prompt was followed by \textbackslash\textbackslash\footnote{which were meant to be newline characters, but remained backslashes because of a bug, which did not hinder GPT-3 from generating good texts!} and a starting phrase, which was one of the 4 prompt kinds, mentioned in Section \ref{sec:human-text-sel}. The exact starting phrases are that we have used for each category are provided in the technical note and in the file \emph{prompts.yaml} in the associated Github repository (\url{https://github.com/dkalpakchi/SweCTRL-Mini/blob/main/human_eval/prompts.yaml}). Note that in SweCTRL-Mini starting phrases were prompts themselves and have followed directly after the corresponding OCC without any kind of GPT-like initial prompt.

\end{appendices}


\bibliography{sn-bibliography}% common bib file

\end{document}